\pdfoutput=1
\documentclass[11pt]{article}
\usepackage[T1]{fontenc}
\usepackage[utf8]{inputenc}
\usepackage{times}
\usepackage[letterpaper,margin=1in]{geometry}
\usepackage[round,authoryear]{natbib}
\usepackage{amsmath,amssymb}
\usepackage{booktabs}
\usepackage{array}
\newcolumntype{L}[1]{>{\raggedright\arraybackslash}p{#1}}
\usepackage{graphicx}
\usepackage{xcolor}
\usepackage{float}
\usepackage{url}
\usepackage{upquote}
\usepackage[hidelinks]{hyperref}
\hypersetup{pdftitle={Behavioral History Outperforms Descriptions of the Person for LLM Synthetic Personas},
  pdfauthor={Khashayar Pourtaheri, Ahmad Zareei}}

\title{Behavioral History Outperforms Descriptions of the Person\\ for LLM Synthetic Personas}
\author{Khashayar Pourtaheri\\
University of Washington\\
\texttt{pk1373@uw.edu}
\and
Ahmad Zareei\\
\texttt{zareei@berkeley.edu}}
\date{}

\begin{document}
\maketitle

\begin{abstract}
Researchers increasingly use large language models as synthetic personas of survey respondents, to cut the
costs of survey research, market analysis, and policy experimentation. Their validity as substitutes for
particular respondents depends on whether they reproduce individuals' decisions, and behavioral biases shape
many of those decisions. We examine what information helps synthetic respondents predict each individual's
later choices. We compare five conditions with progressively richer information: no personal information,
demographics, personality traits, cognitive scores, and the respondent's earlier survey choices as behavioral
history. We use a two-wave panel of 845 US adults measured on 14 behavioral biases, so each respondent's
earlier answers provide a human test--retest benchmark; behavioral history excludes all items that score the
bias being predicted. At the population level, every condition is close to the human average number of biases
per respondent (7.1--8.1, against 7.1). This aggregate similarity hides differences in variance: persona
descriptions recover only 53--67\% of human between-person variation, whereas adding behavioral history
restores it to about the human level. At the individual level, measured by informedness, persona descriptions
reach only 7--12\% of the human test--retest benchmark, and adding behavioral history raises this to 28\%.
Behavioral history gives the highest estimated informedness in all 17 demographic groups, while persona
descriptions give little or no information for some. Synthetic responses also show stronger education- and
income-related differences than human responses. For predicting individual choices, what people have done
tells a model more than who they are, so synthetic-respondent studies should collect the most relevant
behavioral history available.
\end{abstract}

\section{Introduction}

Synthetic respondents built from large language models (LLMs) could reduce the cost of survey research, market
analysis, and policy experimentation.\footnote{On these uses, see \citet{korinek2023generative} and
\citet{manning2024automated}.} The validity of such respondents, however, depends on what they are used for: a model
may reproduce a population distribution without reproducing the decisions of the individuals who generated it.
Substituting a model for a particular respondent therefore requires a person-level test. We ask which type of
information about a person improves such a model's prediction of that person's later choices, and by how much. We find that
a respondent's past answers improve prediction more than a description of who they are. Measured by informedness,
which is 0, on average, for a predictor that ignores the person, persona descriptions reach only 7--12\% of a human
test--retest benchmark, and adding past answers raises this share to 28\%.

Behavioral-economics surveys are a demanding place to run this test. They measure behavioral biases, which are
systematic departures from standard models of rational choice and accurate beliefs, such as favoring immediate
rewards (present bias), weighting losses more than gains (loss aversion), or overrating one's own ability
(overconfidence) \citep{stango2023behavioral}. We use ``bias'' only in this sense, not in its statistical or
fairness sense. An LLM may give the normatively correct answer while failing to reproduce the target respondent's
choice.\footnote{See \citet{liu2025rational}.} Because these biases shape people's decision-making, a model that
gets the answers right, or matches the population on average, cannot stand in for a particular person unless it also
reproduces that person's biases.

Our evaluation uses the two-wave panel of \citet{stango2023behavioral}, in which 845 US adults completed measures of
14 behavioral biases in 2014--15 and again in 2017. For each respondent, we build a synthetic respondent from that
person's own survey record and ask the model to answer each item as that person would. The model is told more about
the person in each of five nested conditions. In A1 it is told nothing. A2 adds demographics, A3 personality traits,
and A4 cognitive scores. A5 adds the person's 2014--15 answers to other survey items as behavioral history, minus
every item that scores the bias being predicted (the target bias), so the model never directly sees the person's
previous result on that bias. Since nothing else changes between conditions, comparing two adjacent conditions shows what one
more block of information adds to the blocks already in the prompt. As a human benchmark, we ask how well each
person's own 2014--15 answers predict their 2017 target responses. Raw accuracy is a poor measure here. Bias
prevalence runs from 0.17 to 0.85, so a model that always gives the more common response scores well without knowing
anything about the respondent. We instead use informedness, the true-positive rate plus the true-negative rate minus
one, which gives no credit for merely following the more common response.

Behavioral history improves both individual-level prediction and the recovery of between-person variation. Without
any information about the person (A1), the model's informedness is essentially zero. With descriptions of the person
(A2--A4), the model recovers only 53--67\% of the human between-person variation, and some demographic groups get little or
no information from these descriptions. Adding the person's own earlier answers (A5) brings the variation to
approximately the human level and gives the highest informedness of any condition in each of the 17 demographic
groups. None of this shows up in population averages, because every condition gives 7.1--8.1 biases per respondent
on average, close to the human 7.1.

Synthetic bias counts are also more strongly structured by education and income than human bias counts, and
behavioral history does little to change that. Under A2, education and income account for 74\% of the variation
explained. The human figure is 18\%, and in the human data cognitive scores are the largest contributor. Between A4
and A5 the education and income contributions even rise slightly, from 19.8 to 21.1 points of $R^2$, against 2.9 for
humans.

Our main findings suggest that synthetic-respondent studies should give the model the most relevant behavioral
history available for each person, and that they should validate person-level agreement before treating model
responses as substitutes for individual respondents. The behavioral history in our design is limited to answers from
a single earlier survey, which leaves room for future work to test how much richer or more recent behavioral records
improve the prediction of a person's decisions. Section~\ref{sec:limits} discusses the other limits of our design.

Our paper adds to the studies that build LLM agents for real people and test them against those same people, by
measuring how much each kind of personal information improves the prediction of a person's choices.
\citet{park2024generative} and \citet{jia2026personas} ground such agents in interviews, surveys, or longitudinal
histories and compare the results. \citet{toubia2025twin} find that richer personal records can improve prediction,
although substantial individual-level errors remain.\footnote{Other person-matched studies include
\citet{choi2025pwyw}, \citet{kinzinger2026synthetic}, and \citet{garzon2026demographics}.} Measuring such gains
takes care. \citet{ahn2026itemmean} show that apparent person-level accuracy can largely reflect item means instead
of individual deviations, and related studies find that simple statistical baselines often do better.\footnote{See
also \citet{peng2026funhouse}, \citet{chen2026synthetic}, \citet{wang2026substitutability}, and
\citet{lukauskas2026plausible}.} It has remained unresolved which kind of information about a person (demographics,
personality, cognitive scores, or past answers) best helps a model reproduce that person's responses, how much each
block of information improves the prediction, and whether it restores the real variation across people. We answer
these questions by adding one block at a time for the same respondents and the same 14 behavioral biases.

More broadly, the paper addresses a puzzle in the literature on LLM simulation, where the same models look human in
some studies and fail in others. Many studies find that LLMs reproduce some subgroup distributions, experimental
regularities, and aggregate patterns of economic behavior,\footnote{See \citet{argyle2023out},
\citet{aher2023using}, \citet{filippas2024homo}, \citet{yan2025arithmetic}, \citet{bini2026behavioral}, and
\citet{ashokkumar2026predict}; \citet{hu2026simbench} evaluate group-level simulation.} while others find that
persona prompting compresses human variation, distorts demographic relationships, or introduces prompt artifacts and
identity-specific reasoning errors.\footnote{See \citet{bisbee2024synthetic}, \citet{dominguez2024questioning},
\citet{gupta2024bias}, \citet{hu2024personaeffect}, \citet{park2024diminished}, \citet{gao2025caution},
\citet{wang2025misportray}, and \citet{gui2025challenge}. Appendix~\ref{app:related} reviews further related work.}
Our results show how both findings can hold at once, because agreement in aggregate says little about agreement for
each person. Studies of personas and economic risk-taking find that personas shift risk and cognitive-bias decisions
\citep{iwamoto2026risk,liu2025riskpersona,yoshikawa2026persona}, but they do not test the later choices of the
specific people described, which is the comparison our design makes.

\section{Experimental setup}
\label{sec:setup}

\subsection{Participants and outcomes}

The data are the two-wave panel of \citet{stango2023behavioral}, which follows 845 adults from a US national panel.
They took the same behavioral-economics battery twice, in 2014--15 (wave 1) and in 2017 (wave 2). We always predict
wave 2. Wave 1 has two roles, as the human test--retest benchmark and as the source of the behavioral history used
in A5.

The battery measures 17 biases. We leave out three of them, two that use a graphical budget game and one that needs
a next-day login. Our outcomes are the other 14. The 14 span present bias, inconsistent choice, risk,
overconfidence, mathematical and statistical reasoning, and attention. Each bias is scored for each respondent as a
binary verdict: 1 if the answers show the bias, 0 if they do not. Appendix~\ref{app:biases} lists every instrument
and scoring rule.

\subsection{Persona information and administration}

The model predicts these verdicts from a persona built from the respondent's own survey record: wave-2 descriptions
of the person and, in A5, wave-1 answers. Table~\ref{tab:conditions} lists the five nested conditions. The
demographic block in A2 has 16 fields, such as age, gender, race, Hispanic origin, education, and income. A3 adds
the Big Five scores from the BFI-10 \citep{rammstedt2007bfi10}, which the panel administered. A4 adds four short
tests from the same survey (numeracy, financial literacy, a number-series test of fluid intelligence, and a timed
color--word test of executive function).

\begin{table}[!htb]
\caption{\textbf{Five nested prompt conditions.} Each condition adds one information block to the
previous condition. A5 includes respondent-specific behavioral evidence, but the target bias block
is withheld.}
\label{tab:conditions}
\centering\small
\begin{tabular}{@{}llp{7.4cm}@{}}
\toprule
& Condition & Prompt information \\
\midrule
A1 & No identity        & the survey question alone \\
A2 & Demographics       & adds 16 wave-2 demographic fields \\
A3 & Personality        & adds the BFI-10 Big Five, with scale notes \\
A4 & Cognitive          & adds four cognitive scores, each with its scale \\
A5 & Behavioral history & adds the person's wave-1 responses, excluding all items used to score the target bias \\
\bottomrule
\end{tabular}
\end{table}

A5 adds behavioral history: the respondent's wave-1 answers to other survey items. These answers come from the same
battery as the outcomes, so we have to keep the target out of the prompt. We therefore remove the whole item block
that scores the target bias, so the model never sees the person's earlier answers to the bias it predicts. Answers
to related items stay in, and they may carry information about correlated behavior. Because the conditions are
nested, results we attribute to behavioral history refer to A5, which adds earlier answers to the full persona
description.

Each item goes to the model in a fresh context. There is no conversation history. Items that must be answered
together are the one exception (for example, a construct defined only under joint evaluation). Keeping items apart
copies the module structure of the original survey and avoids artifacts from accumulated context
\citep{dominguez2024questioning}. If an item refers to an earlier answer, the prompt restates the answer the model
gave in the same condition. We generated all responses with Claude Opus 4.8 in August 2026. Each respondent, item,
and condition got one generation. Appendix~\ref{app:prompts} documents the prompts, withholding rules, and
administration.

\subsection{Scoring and evaluation}

The model's answers and the human answers are scored by one program that follows the source study's published rules.
No outcome is rescaled. The main outcome is each respondent's raw wave-2 verdict. We do not correct either side for
measurement error. Individual-level agreement is measured with informedness, $J=\mathrm{TPR}+\mathrm{TNR}-1$. The
true-positive rate (TPR) is the share of respondents with the bias whose verdict is predicted correctly. The
true-negative rate (TNR) is the same share for respondents without it. $J$ is 1 for perfect prediction. It is 0 on
average for a predictor that ignores the person, such as one that always gives the same answer or guesses at random.
We average $J$ over the 14 biases and report each condition as a share of the human test--retest value
(Appendix~\ref{app:estimators}).

Informedness is one of several measures, which we group by the question they answer. We assess population
resemblance with bias counts, a variance decomposition of the person-by-bias matrix, and a respondent-grouped
classifier that distinguishes synthetic from human profiles. We assess individual fidelity with per-bias agreement,
person-by-bias residual correlation, and an identification AUC. This AUC asks whether a synthetic profile of continuous
bias scores is closer to its own respondent than to another respondent. For both AUCs, 0.5 is chance. A classifier
AUC near 0.5 means synthetic and human profiles are hard to tell apart. A higher identification AUC means synthetic
profiles are closer to their own respondent.

These measures do not all use the same respondents, because each needs different observed cells. The bias-count
distribution uses 696 respondents with all 14 biases in every condition. The variance decomposition uses the 518 who
also have a complete matrix for the human target. Per-bias prevalence uses all available cases. Per-bias agreement
uses the common retest sample (the target, the human benchmark, and every condition all observed). The profile
analyses use 829 respondents with at least 7 jointly observed biases. Appendix~\ref{app:samples} gives the full
inclusion rules.

\section{Results}
\label{sec:results}

\subsection{Aggregate counts conceal a different population}
\label{sec:population}

\begin{figure}[!htb]
\centering
\includegraphics[width=\linewidth]{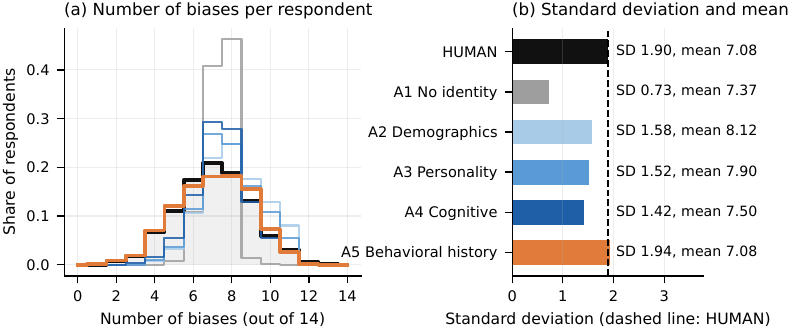}
\caption{\textbf{How many of the 14 biases each respondent shows, and how the distribution differs
by condition.} Panel (a) shows the bias-count distribution, and panel (b) reports each condition's standard
deviation and mean; colors are the same in both panels. $n=696$ respondents with all 14 biases
available in every condition.}
\label{fig:bcount}
\end{figure}

The average number of biases per respondent is close to the human average in all five conditions, but A1--A4 show less variation across respondents than humans do (Figure~\ref{fig:bcount}). Humans show 7.08 biases on average ($\mathrm{SD}=1.90$), and A5 has the same mean and a similar standard deviation ($\mathrm{SD}=1.94$). A1 is the extreme case. Its SD is only 0.73, and it places 87\% of respondents in two adjacent count bins. A2--A4 lie between A1 and humans, so only A5 comes close to humans on both the average and the spread.

The persona descriptions (A2--A4) recover only 53--67\% of the human between-person variation, while behavioral history restores approximately the human amount (Table~\ref{tab:variance}). We split the total variation in the person-by-bias matrix into differences between biases, differences between people, and a residual specific to each person--bias pair, and report each part as its share of the total sum of squares. In A1, differences between biases explain 72.8 times as much variation as differences between people. The ratio is between 12.2 and 15.7 in A2--A4 and 2.6 in humans. With a ratio of 5.1 and a larger residual share than A1--A4, A5 is the closest of the synthetic conditions. Still, a Mantel test \citep{mantel1967detection}, run without one mechanically confounded pair of biases (Appendix~\ref{app:limits}), gives no evidence that any condition recovers the human pattern of relationships among biases. Synthetic profiles are also easy to tell apart from human ones. On the primary sample ($n=829$), a classifier with folds grouped by respondent separates the two with AUC 1.000 under A1 and 0.976 under A5, yet it cannot separate the two human waves (AUC 0.476; 0.510 on complete profiles, Table~\ref{tab:complete-profile}).

\begin{table}[!htb]
\caption{\textbf{Variance decomposition of the person-by-bias matrix.} The complete
$n\times B$ binary matrix is separated into between-bias, between-person, and residual shares of
total variance (here $n=518$ respondents and $B=14$ biases).}
\label{tab:variance}
\centering\small\setlength{\tabcolsep}{3.5pt}
\begin{tabular}{@{}lccccc@{}}
\toprule
Condition & Bias & Person & Residual & Person/HUMAN & Bias/Person \\
\midrule
HUMAN (wave 2)      & 0.191 & 0.074 & 0.735 & 1.00 & 2.6\\
\midrule
A1 No identity      & 0.834 & 0.011 & 0.155 & 0.15 & 72.8\\
A2 Demographics     & 0.605 & 0.049 & 0.345 & 0.67 & 12.2\\
A3 Personality      & 0.567 & 0.046 & 0.386 & 0.62 & 12.3\\
A4 Cognitive        & 0.614 & 0.039 & 0.347 & 0.53 & 15.7\\
A5 Behavioral history & 0.395 & 0.078 & 0.527 & 1.05 & 5.1\\
\bottomrule
\end{tabular}
\end{table}

\begin{figure}[!htb]
\centering
\includegraphics[width=\linewidth]{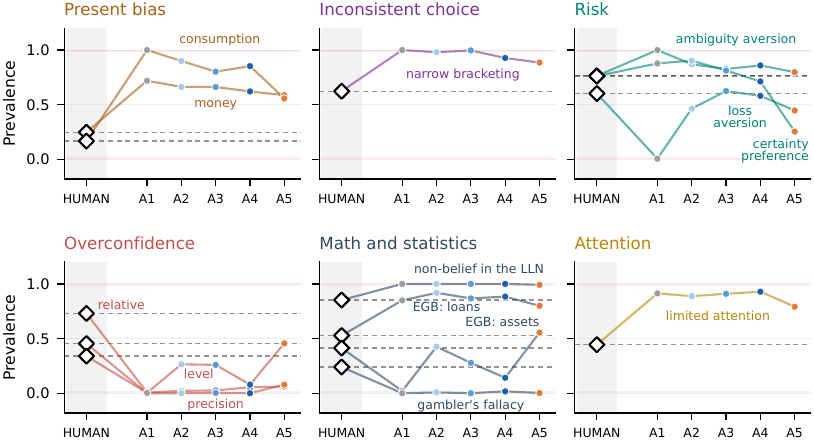}
\caption{\textbf{Prevalence of each bias across the information conditions.} Each panel shows
one bias family: HUMAN (gray column), then A1--A5. Dashed lines mark each bias's human wave-2
prevalence, so the vertical distance from a dashed line is the aggregate error. Shaded bands mark
prevalence near zero or one. Each bias uses its available cases ($n=629$--845). EGB is
exponential-growth bias and LLN the law of large numbers.}
\label{fig:prevladder}
\end{figure}

In per-bias prevalence (Figure~\ref{fig:prevladder}; values in Table~\ref{tab:prev-values}), A1 gives rates close
to zero or one for most biases. A2--A5 vary more, but they still differ substantially from
humans. In every model condition the gambler's fallacy almost never appears, and non-belief in the law of large
numbers appears in almost everyone. A5 is closest to human prevalence on 9 of the 14 biases, while other
conditions come closer on loss aversion, certainty preference, overconfidence-level, gambler's fallacy, and
exponential-growth bias (EGB) on assets. Matching the average number of biases does not mean that
synthetic respondents reproduce the human population, because some biases can be overpredicted and others
underpredicted, and these errors can offset each other.

\subsection{Behavioral history gives the strongest individual-level signal}
\label{sec:ladder}

Raw accuracy barely moves across conditions, from 0.550 in A1 to 0.569 in A5, against 0.709 for the human retest
(Table~\ref{tab:accuracy}). Informedness separates the conditions much more clearly. A1 has essentially none
($J=0.001$). A2--A4 reach only $J=0.020$--$0.033$, or 7--12\% of the human benchmark. A5 reaches $J=0.078$, or 28\%,
so behavioral history provides the strongest individual-level signal, although it remains far below the human
benchmark. It also makes the predictions more balanced. The gap $|\mathrm{TPR}-\mathrm{TNR}|$, which is 0 when
respondents with and without the bias are predicted equally well, falls from 0.90 in A1 to 0.53 in A5, against 0.26
for the human benchmark.

\begin{table}[!htb]
\caption{\textbf{Behavioral history improves individual-level prediction.} Each synthetic verdict is compared with the same respondent's observed wave-2 verdict, and results are averaged across the 14 biases on each bias's common retest sample ($n=526$--843). HUMAN compares the same person's wave-1 and wave-2 verdicts.}
\label{tab:accuracy}
\vspace{4pt}
\centering\footnotesize\setlength{\tabcolsep}{2.5pt}
\begin{tabular}{@{}lcccccc@{}}
\toprule
& \shortstack{HUMAN\\retest}
& \shortstack{A1\\No identity}
& \shortstack{A2\\Demographics}
& \shortstack{A3\\Personality}
& \shortstack{A4\\Cognitive}
& \shortstack{A5\\Behavioral\\history}\\
\midrule
Mean accuracy                         & 0.709 & 0.550 & 0.558 & 0.564 & 0.568 & \textbf{0.569}\\
\addlinespace[5pt]
Mean informedness $J$                 & 0.283 & 0.001 & 0.020 & 0.024 & 0.033 & \textbf{0.078}\\
\quad Share of HUMAN                 & 1.00 & 0.00 & 0.07 & 0.08 & 0.12 & \textbf{0.28}\\
\addlinespace[5pt]
Mean $|\mathrm{TPR}-\mathrm{TNR}|$ & 0.26 & 0.90 & 0.70 & 0.71 & 0.72 & \textbf{0.53}\\
\quad Ratio to HUMAN    & 1.00 & 3.42 & 2.65 & 2.69 & 2.74 & \textbf{2.00}\\
\bottomrule
\end{tabular}
\end{table}

\begin{table}[!htb]
\caption{\textbf{Mean informedness $J$ by demographic group.} We compute
$J=\mathrm{TPR}+\mathrm{TNR}-1$ within each group-by-bias cell and then average across estimable
biases. HUMAN is the group's wave-1--wave-2 benchmark, and $n$ is the median valid sample across
included biases. Other is the source panel's own category, not a merger of small groups. The last column is A5 minus
A4; bold values exceed the mean gain across groups (0.039).}
\label{tab:groupinf}
\centering\footnotesize\setlength{\tabcolsep}{2.5pt}
\begin{tabular}{@{}llrrrrrrrr@{}}
\toprule
Attribute & Group & $n$
& \shortstack{HUMAN\\{\scriptsize retest}}
& \shortstack{A1\\{\scriptsize No identity}}
& \shortstack{A2\\{\scriptsize Demographics}}
& \shortstack{A3\\{\scriptsize Personality}}
& \shortstack{A4\\{\scriptsize Cognitive}}
& \shortstack{A5\\{\scriptsize Behavioral}\\{\scriptsize history}}
& \shortstack{A5$-$A4}\\
\midrule
income & low & 308 & 0.269 & 0.010 & 0.018 & 0.020 & 0.025 & 0.071 & \textbf{0.047}\\
 & mid & 308 & 0.285 & $-0.012$ & $-0.002$ & 0.003 & 0.037 & 0.089 & \textbf{0.052}\\
 & high & 192 & 0.263 & 0.010 & 0.019 & 0.027 & 0.027 & 0.030 & 0.003\\
\midrule
education & high school or less & 129 & 0.255 & 0.001 & 0.001 & 0.020 & 0.047 & 0.057 & 0.010\\
 & some college & 289 & 0.283 & $-0.001$ & 0.002 & 0.005 & 0.022 & 0.059 & 0.037\\
 & bachelor's & 223 & 0.281 & $-0.005$ & 0.006 & 0.023 & 0.039 & 0.087 & \textbf{0.048}\\
 & postgraduate & 165 & 0.275 & 0.011 & 0.007 & $-0.005$ & 0.014 & 0.059 & \textbf{0.045}\\
\midrule
ethnicity & Black & 86 & 0.206 & 0.017 & $-0.002$ & 0.040 & 0.016 & 0.045 & 0.029\\
 & Other & 55 & 0.267 & $-0.031$ & 0.046 & 0.048 & 0.035 & 0.116 & \textbf{0.080}\\
 & White & 637 & 0.290 & 0.000 & 0.023 & 0.026 & 0.035 & 0.077 & \textbf{0.042}\\
\midrule
age & young & 279 & 0.254 & 0.007 & 0.042 & 0.033 & 0.024 & 0.055 & 0.031\\
 & middle & 285 & 0.282 & $-0.003$ & 0.007 & 0.016 & 0.031 & 0.100 & \textbf{0.069}\\
 & older & 246 & 0.306 & $-0.001$ & 0.004 & 0.016 & 0.049 & 0.071 & 0.022\\
\midrule
gender & Female & 514 & 0.275 & 0.001 & 0.016 & 0.016 & 0.032 & 0.080 & \textbf{0.048}\\
 & Male & 297 & 0.284 & 0.003 & 0.023 & 0.034 & 0.030 & 0.064 & 0.034\\
\midrule
origin & Hispanic & 105 & 0.212 & $-0.027$ & 0.054 & 0.028 & 0.053 & 0.065 & 0.013\\
 & not Hispanic & 705 & 0.290 & 0.005 & 0.013 & 0.022 & 0.030 & 0.079 & \textbf{0.049}\\
\bottomrule
\end{tabular}
\end{table}

\begin{figure}[!htb]
\centering
\includegraphics[width=0.8\linewidth]{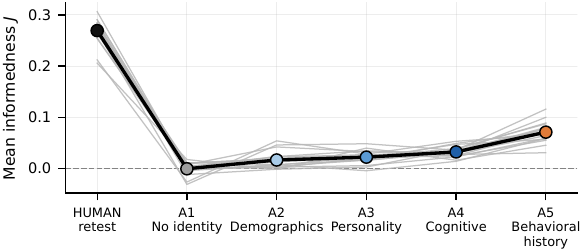}
\caption{\textbf{Behavioral history yields the highest informedness across demographic groups.}
  Each thin line shows mean informedness $J$ for one of the 17 demographic groups, and the bold line shows the average across groups. HUMAN retest is each group's wave-1--wave-2 benchmark; the dashed line marks $J=0$.}
\label{fig:ladder}
\end{figure}

The same ordering holds within demographic groups. A5 has positive informedness, and the highest point estimate of
any model condition, in all 17 groups (Table~\ref{tab:groupinf}). The description-based conditions give values near
or below zero for some groups. If the 17 groups are weighted equally, informedness rises from about zero in A1 to
0.016 in A2, 0.022 in A3, 0.032 in A4, and 0.071 in A5. A5 thus reaches 26\% of the human group-average benchmark of 0.269
(Figure~\ref{fig:ladder}).

Behavioral history helps some groups more than others. Relative to A4, A5 raises informedness by 0.039
on average across the 17 groups, and by more than this in 9 of them (bold in Table~\ref{tab:groupinf}),
including postgraduate respondents, for whom A3 and A4 give the lowest informedness of any group
($-0.005$ and $0.014$). The gain is smallest for high-income (0.003), high-school-or-less (0.010), and Hispanic
(0.013) respondents. The human benchmark itself also differs across groups, from 0.206 for Black and
0.212 for Hispanic respondents to 0.306 for older respondents. Measured against its own benchmark, A5
reaches 21--35\% in 15 of 17 groups, and more for Hispanic (31\%) than non-Hispanic respondents (27\%);
the exceptions are high-income (12\%) and Other respondents (43\%, $n=55$). A5 also begins to rank groups the way humans do
(rank correlation 0.49 with HUMAN, against 0.20 for A4).

A5 has the highest informedness in 104 of the 207 bias-by-subgroup cells we can estimate (50.2\%; Table~\ref{tab:grid}). Crossing 14 biases with 15 subgroup levels, the groups of Table~\ref{tab:groupinf} without the two Hispanic-origin levels, gives 210 cells. We drop three cells, two with fewer than 40 valid observations and one with fewer than five observations in one outcome class. Even the best synthetic condition stays below the human benchmark in 173 of the 207 cells (84\%).

\begin{table}[!htb]
\caption{\textbf{Best- and worst-performing conditions by bias and demographic subgroup.}
  For each of the 207 estimable bias-by-subgroup cells, A1--A5 are ranked by informedness $J$. ``Best in'' and ``worst in'' report
  how often each condition has the highest or lowest point estimate.}
\label{tab:grid}
\centering\footnotesize\setlength{\tabcolsep}{6pt}
\begin{tabular}{@{}lrrrr@{}}
\toprule
Condition & Best in & \% & Worst in & \% \\
\midrule
A1 No identity                & 22 & 10.6\% & \textbf{90} & \textbf{43.5\%} \\
A2 Demographics               & 19 &  9.2\% & 48 & 23.2\% \\
A3 Personality                & 19 &  9.2\% & 25 & 12.1\% \\
A4 Cognitive                  & 43 & 20.8\% & 24 & 11.6\% \\
A5 Behavioral history         & \textbf{104} & \textbf{50.2\%} & 20 & 9.7\% \\
\bottomrule
\end{tabular}
\end{table}

Checks on whole profiles point the same way (Appendix~\ref{app:extra}). After person and bias means are removed,
A5's residual correlation with human responses on the primary sample ($n=829$) is 0.070. That is 24\% of the human
test--retest value of 0.287; A4 reaches 0.023, or 8\% (Table~\ref{tab:complete-profile}). Identification is harder.
A5's identification AUC is only 0.552, against 0.718 for the human retest ($n=828$;
Table~\ref{tab:primary-identification}). When we reassign A5 outputs among demographically matched respondents,
informedness falls from 0.078 to 0.026 ($p<0.001$). The gain thus depends on matching each output to the intended
respondent, although this test does not isolate why behavioral history helps. The gain also holds on continuous scores
(Appendix~\ref{app:continuous}) and varies across respondents (Appendix~\ref{app:person-gains}).

\subsection{Persona conditions reshape demographic disparities}
\label{sec:whom}

\begin{figure}[!htb]
\centering
\includegraphics[width=\linewidth]{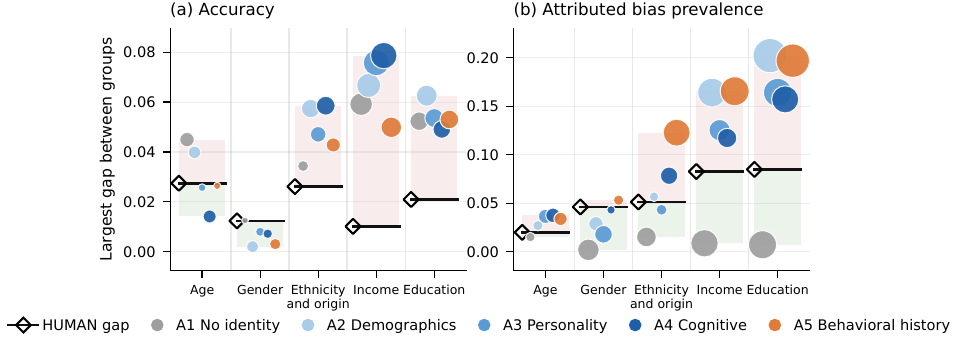}
\caption{\textbf{Persona conditions reshape demographic disparities relative to humans.} Each point shows the largest gap across an attribute's groups in prediction accuracy (a) or attributed bias prevalence (b); the black line marks the human gap. Red shading marks gaps larger than the human gap and green shading smaller ones, and larger markers lie farther from the human gap.}
\label{fig:columns}
\end{figure}

Persona conditioning widens some demographic gaps and narrows others (Figure~\ref{fig:columns}; values in
Table~\ref{tab:gap-values}). We compare two kinds of gap: gaps between groups in prediction accuracy, and gaps in
the prevalence of biases attributed to each group. Income and education gaps are larger than the
human gaps in accuracy under every condition, and in attributed prevalence under every condition with a persona
profile (A2--A5). Age and gender do not follow this pattern. The gender gap shrinks almost everywhere. Every
condition with a persona profile has a smaller gender gap in accuracy than humans, and in attributed prevalence
only A4 comes close to the human gap (0.043 versus 0.046) and only A5 is above it (0.053). For age, richer
conditions narrow the accuracy gap, but A2--A5 widen the gap in attributed prevalence. The conditions can also
reverse a human pattern. Among human respondents, loss aversion rises modestly with income, but it falls with
income under every condition with a persona profile (Appendix~\ref{app:lossaversion}).

\begin{figure}[!htb]
\centering
\includegraphics[width=\linewidth]{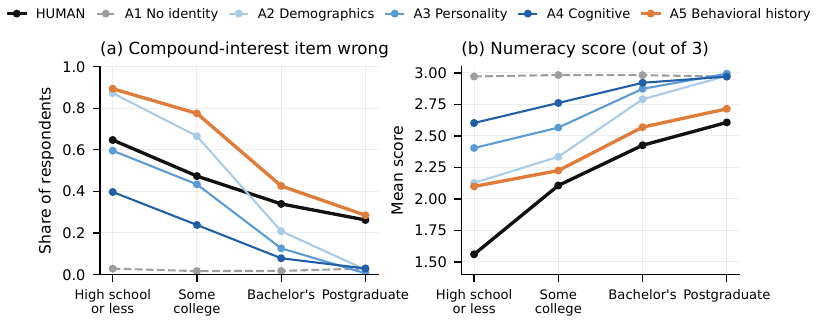}
\caption{\textbf{Synthetic respondents misrepresent how skill varies with education.} (a) Share answering the compound-interest item wrongly. (b) Average number of the three numeracy questions answered correctly (0--3).}
\label{fig:incap}
\end{figure}

One item shows the gap between solving a question and answering it as the respondent would. On the compound-interest
item (Figure~\ref{fig:incap}; values in Table~\ref{tab:incap-values}), A1 gets the answer right for almost every
respondent, regardless of education. A2, A3, and A5 produce a steeper education gradient than humans show, and
A2--A4 make postgraduate personas almost always correct. A5 comes close to the human error rate for postgraduates
(28\% versus 26\%) but predicts too many errors in the other education groups. On this item, behavioral history does
not recover the human pattern.

Numeracy, in panel (b), shows a different distortion. Every condition answers more questions correctly than the
respondents it represents, most of all for respondents with a high school education or less (1.56 correct for
humans, against 2.10--2.97 for the conditions). For postgraduate personas, A1--A4 answer almost every question
correctly (2.97--2.99, against 2.61 for humans). A5 stays closer (2.72) and is the closest condition in every
education group.

\section{Discussion}
\label{sec:discussion}

Behavioral history changes which respondent attributes explain synthetic bias counts, but not the weight of
education and income (Table~\ref{tab:blockcoef}). Adding past answers to A4 lowers the variation explained by
respondent attributes from 47.5\% to 42.9\%. The fall comes mostly from smaller cognitive, personality, and age
contributions. The education and income contributions do not shrink. They move from 19.8 to 21.1 points of $R^2$,
against 2.9 for humans, so their share of the explained variation rises from 42\% under A4 to 49\% under A5, far
above the human 18\%. Our behavioral history is a single earlier survey. Richer or more recent records may weaken
this link, and we see that as the main direction for future work.

\begin{table}[!htb]
\caption{\textbf{Decomposition of explained variation in human and synthetic bias counts.}
Cells are Shapley contributions to $R^2$ (percentage points) from regressing each column's bias count on
respondent attributes (757 respondents; \citealp{shapley1953value}). Blocks absent from a condition's prompt (the Big Five and cognitive scores for A2; the cognitive
scores for A3) measure correlation with the synthetic answers, not use of that information. Stars mark jointly significant blocks (Wald test with Benjamini--Hochberg correction,
\citealp{benjamini1995controlling}; $^{*}q<.05$, $^{**}q<.01$, $^{***}q<.001$); every condition's
education-and-income share exceeds the human share ($q<.001$).}
\label{tab:blockcoef}
\centering\footnotesize\setlength{\tabcolsep}{3.5pt}
\begin{tabular}{@{}lccccc@{}}
\toprule
Construct
& \shortstack{HUMAN\\{\scriptsize (wave 2)}}
& \shortstack{A2\\{\scriptsize Demographics}}
& \shortstack{A3\\{\scriptsize Personality}}
& \shortstack{A4\\{\scriptsize Cognitive}}
& \shortstack{A5\\{\scriptsize Behavioral}\\{\scriptsize history}}\\
\midrule
Age & $0.09$ & $0.21$ & $0.34$ & $1.01$ & $0.12$\\
Gender & $\textbf{1.56}^{*}$ & $\textbf{0.87}^{*}$ & $0.45$ & $\textbf{1.65}^{*}$ & $\textbf{1.71}^{*}$\\
Ethnicity \& Hispanic origin & $0.84$ & $0.59$ & $0.47$ & $1.10$ & $1.33$\\
Education & $1.64$ & $\textbf{22.69}^{***}$ & $\textbf{16.91}^{***}$ & $\textbf{12.53}^{***}$ & $\textbf{13.80}^{***}$\\
Income & $1.29$ & $\textbf{11.31}^{***}$ & $\textbf{8.89}^{***}$ & $\textbf{7.25}^{**}$ & $\textbf{7.26}^{**}$\\
Big Five (5 traits) & $\textbf{1.93}^{*}$ & $0.80$ & $\textbf{3.81}^{***}$ & $\textbf{3.34}^{***}$ & $1.44$\\
Cognitive scores (4) & $\textbf{8.65}^{***}$ & $\textbf{8.15}^{*}$ & $4.82$ & $\textbf{20.12}^{***}$ & $\textbf{15.91}^{***}$\\
\midrule
$R^2$ from the seven blocks & 16.0 & 44.6 & 35.7 & 47.0 & 41.6\\
Total $R^2$ (18 attributes; App.~\ref{app:samples}) & 16.5 & 46.2 & 37.2 & 47.5 & 42.9\\
Education and income, \% of total $R^2$ & 18\% & 74\% & 69\% & 42\% & 49\%\\
\bottomrule
\end{tabular}
\end{table}

\section{Limitations}
\label{sec:limits}

We study one model, one US panel, and 14 biases measured in text. The findings may not hold in other settings. We
also generated each response only once per respondent, item, and condition, so part of the measured between-person
variation may be run-to-run randomness. Finally, the public data may have been in the model's training data. If so,
our design cannot tell learned population rates apart from memorized individual records
\citep{ludwig2026econometric,lopezlira2025memorization}.

\section{Conclusion}
\label{sec:conclusion}

Resembling the population in aggregate does not show that an LLM can stand in for an individual respondent.
Descriptions of a person's demographics, personality traits, and cognitive scores give plausible population
averages. Yet they carry little information about the specific person, and they can distort the differences between
groups. The person's own past answers do better, restoring more of the between-person variation and improving
individual-level prediction, although most of the human test--retest signal is still missing. The evidence thus
points synthetic-respondent studies toward behavioral data, and toward validating person-level agreement before model
responses replace individual respondents.

\subsection*{Reproducibility statement}

The appendices report the bias definitions, prompt protocol, scoring rules, sample construction, estimators,
continuous-score checks, and coverage limits. Human and model responses are scored with the same rules, and each
analysis states its sample in the text, a caption, or the appendix. The fixed model responses and the analysis code
are available from the authors on request. The de-identified source survey data are available from the original
study's public replication package.

\subsection*{Ethics statement}

This study uses de-identified secondary data from a published replication package and collects no new
human-subject data. The personas contain structured survey fields without names or free text, and the
target outcome is withheld from every prediction prompt. The demographic analyses describe a research
simulation and should not be read as evidence about a deployed decision system.

\subsection*{AI use statement}

Large language models were used in several ways in this work. Most centrally, Claude Opus 4.8
generated the synthetic survey responses that the paper studies (Appendix~\ref{app:prompts}). We also
used LLM-based tools to help search the related literature, to write and check analysis code, and to
edit the writing and format the manuscript. The authors checked every cited work against its source, verified the
code and all reported numbers, and revised and approved all text, and they take full responsibility
for the content of the paper.

\bibliographystyle{plainnat}
\bibliography{references}

\begin{thebibliography}{59}
\providecommand{\natexlab}[1]{#1}
\providecommand{\url}[1]{\texttt{#1}}
\expandafter\ifx\csname urlstyle\endcsname\relax
  \providecommand{\doi}[1]{doi: #1}\else
  \providecommand{\doi}{doi: \begingroup \urlstyle{rm}\Url}\fi

\bibitem[Aher et~al.(2023)Aher, Arriaga, and Kalai]{aher2023using}
Gati Aher, Rosa~I. Arriaga, and Adam~Tauman Kalai.
\newblock Using large language models to simulate multiple humans and replicate
  human subject studies.
\newblock In \emph{Proceedings of the 40th International Conference on Machine
  Learning}, volume 202 of \emph{Proceedings of Machine Learning Research},
  pages 337--371, 2023.

\bibitem[Ahn et~al.(2026)Ahn, Mao, and Lee]{ahn2026itemmean}
Daehwan Ahn, Chengfeng Mao, and Dokyun Lee.
\newblock Item-mean surrogates: Why richer persona data fail to improve {LLMs}
  as human surrogates.
\newblock \emph{arXiv preprint arXiv:2608.29455}, 2026.

\bibitem[Argyle et~al.(2023)Argyle, Busby, Fulda, Gubler, Rytting, and
  Wingate]{argyle2023out}
Lisa~P. Argyle, Ethan~C. Busby, Nancy Fulda, Joshua~R. Gubler, Christopher
  Rytting, and David Wingate.
\newblock Out of one, many: Using language models to simulate human samples.
\newblock \emph{Political Analysis}, 31\penalty0 (3):\penalty0 337--351, 2023.

\bibitem[Ashokkumar et~al.(2026)Ashokkumar, Hewitt, Ghezae, and
  Willer]{ashokkumar2026predict}
Ashwini Ashokkumar, Luke Hewitt, Isaias Ghezae, and Robb Willer.
\newblock Large language models can predict the results of social science
  experiments.
\newblock \emph{Nature}, 656\penalty0 (8126):\penalty0 115--122, 2026.

\bibitem[Benjamini and Hochberg(1995)]{benjamini1995controlling}
Yoav Benjamini and Yosef Hochberg.
\newblock Controlling the false discovery rate: A practical and powerful
  approach to multiple testing.
\newblock \emph{Journal of the Royal Statistical Society: Series B
  (Methodological)}, 57\penalty0 (1):\penalty0 289--300, 1995.

\bibitem[Bini et~al.(2026)Bini, Cong, Huang, and Jin]{bini2026behavioral}
Pietro Bini, Lin~William Cong, Xing Huang, and Lawrence~J. Jin.
\newblock Behavioral economics of {AI}: {LLM} biases and corrections.
\newblock Working Paper 34745, National Bureau of Economic Research, 2026.

\bibitem[Binz et~al.(2025)Binz, Akata, Bethge, Br{\"a}ndle, Callaway,
  Coda-Forno, Dayan, Demircan, Eckstein, {\'E}ltet{\H{o}}, Griffiths, Haridi,
  Jagadish, Ji-An, Kipnis, Kumar, Ludwig, Mathony, Mattar, Modirshanechi, Nath,
  Peterson, Rmus, Russek, Saanum, Schubert, Schulze~Buschoff, Singhi, Sui,
  Thalmann, Theis, Truong, Udandarao, Voudouris, Wilson, Witte, Wu, Wulff,
  Xiong, and Schulz]{binz2025centaur}
Marcel Binz, Elif Akata, Matthias Bethge, Franziska Br{\"a}ndle, Fred Callaway,
  Julian Coda-Forno, Peter Dayan, Can Demircan, Maria~K. Eckstein, No{\'e}mi
  {\'E}ltet{\H{o}}, Thomas~L. Griffiths, Susanne Haridi, Akshay~K. Jagadish,
  Li~Ji-An, Alexander Kipnis, Sreejan Kumar, Tobias Ludwig, Marvin Mathony,
  Marcelo Mattar, Alireza Modirshanechi, Surabhi~S. Nath, Joshua~C. Peterson,
  Milena Rmus, Evan~M. Russek, Tankred Saanum, Johannes~A. Schubert, Luca~M.
  Schulze~Buschoff, Nishad Singhi, Xin Sui, Mirko Thalmann, Fabian~J. Theis,
  Vuong Truong, Vishaal Udandarao, Konstantinos Voudouris, Robert Wilson,
  Kristin Witte, Shuchen Wu, Dirk~U. Wulff, Huadong Xiong, and Eric Schulz.
\newblock A foundation model to predict and capture human cognition.
\newblock \emph{Nature}, 644\penalty0 (8078):\penalty0 1002--1009, 2025.

\bibitem[Bisbee et~al.(2024)Bisbee, Clinton, Dorff, Kenkel, and
  Larson]{bisbee2024synthetic}
James Bisbee, Joshua~D. Clinton, Cassy Dorff, Brenton Kenkel, and Jennifer~M.
  Larson.
\newblock Synthetic replacements for human survey data? {T}he perils of large
  language models.
\newblock \emph{Political Analysis}, 32\penalty0 (4):\penalty0 401--416, 2024.

\bibitem[Boelaert et~al.(2025)Boelaert, Coavoux, Ollion, Petev, and
  Pr{\"a}g]{boelaert2025machinebias}
Julien Boelaert, Samuel Coavoux, {\'E}tienne Ollion, Ivaylo Petev, and Patrick
  Pr{\"a}g.
\newblock Machine bias. {H}ow do generative language models answer opinion
  polls?
\newblock \emph{Sociological Methods \& Research}, 54\penalty0 (3):\penalty0
  1156--1196, 2025.

\bibitem[Bollen et~al.(2026)Bollen, Higton, and Sands]{bollen2026local}
Paige Bollen, Joe Higton, and Melissa Sands.
\newblock Nationally representative, locally misaligned: The biases of
  generative artificial intelligence in neighborhood perception.
\newblock \emph{Political Analysis}, 34\penalty0 (3):\penalty0 479--487, 2026.

\bibitem[Brucks and Toubia(2025)]{brucks2025prompt}
Melanie Brucks and Olivier Toubia.
\newblock Prompt architecture induces methodological artifacts in large
  language models.
\newblock \emph{PLOS ONE}, 20\penalty0 (4):\penalty0 e0319159, 2025.

\bibitem[Chen et~al.(2023)Chen, Liu, Shan, and Zhong]{chen2023rationality}
Yiting Chen, Tracy~Xiao Liu, You Shan, and Songfa Zhong.
\newblock The emergence of economic rationality of {GPT}.
\newblock \emph{Proceedings of the National Academy of Sciences}, 120\penalty0
  (51):\penalty0 e2316205120, 2023.

\bibitem[Chen et~al.(2026)Chen, Zhu, and Zheng]{chen2026synthetic}
Zihan Chen, Di~Zhu, and Lei~Nico Zheng.
\newblock When synthetic users fail: A cross-domain benchmark of
  {LLM}-simulated human survey responses.
\newblock \emph{arXiv preprint arXiv:2607.26348}, 2026.

\bibitem[Choi et~al.(2025)Choi, Park, Lee, Shin, Jin, and Kim]{choi2025pwyw}
Junhyuk Choi, Hyeonchu Park, Haemin Lee, Hyebeen Shin, Hyun~Joung Jin, and
  Bugeun Kim.
\newblock Pay what {LLM} wants: Can {LLM} simulate economics experiment with
  522 real-human persona?
\newblock \emph{arXiv preprint arXiv:2508.03262}, 2025.

\bibitem[Cruz et~al.(2024)Cruz, Hardt, and Mendler-D{\"u}nner]{cruz2024risk}
Andr{\'e}~F. Cruz, Moritz Hardt, and Celestine Mendler-D{\"u}nner.
\newblock Evaluating language models as risk scores.
\newblock In \emph{Advances in Neural Information Processing Systems},
  volume~37, pages 97378--97407, 2024.

\bibitem[Cui et~al.(2025)Cui, Li, and Zhou]{cui2025replication}
Ziyan Cui, Ning Li, and Huaikang Zhou.
\newblock A large-scale replication of scenario-based experiments in psychology
  and management using large language models.
\newblock \emph{Nature Computational Science}, 5\penalty0 (8):\penalty0
  627--634, 2025.

\bibitem[Dom{\'\i}nguez-Olmedo et~al.(2024)Dom{\'\i}nguez-Olmedo, Hardt, and
  Mendler-D{\"u}nner]{dominguez2024questioning}
Ricardo Dom{\'\i}nguez-Olmedo, Moritz Hardt, and Celestine Mendler-D{\"u}nner.
\newblock Questioning the survey responses of large language models.
\newblock In \emph{Advances in Neural Information Processing Systems},
  volume~37, pages 45850--45878, 2024.

\bibitem[Filippas et~al.(2024)Filippas, Horton, and Manning]{filippas2024homo}
Apostolos Filippas, John~J. Horton, and Benjamin~S. Manning.
\newblock Large language models as simulated economic agents: What can we learn
  from {Homo Silicus}?
\newblock In \emph{Proceedings of the 25th ACM Conference on Economics and
  Computation}, pages 614--615, 2024.

\bibitem[Fr{\"o}hling et~al.(2026)Fr{\"o}hling, Rupprecht, Strohmaier, and
  Wagner]{frohling2026persona}
Leon Fr{\"o}hling, Jens Rupprecht, Markus Strohmaier, and Claudia Wagner.
\newblock When persona attributes improve population alignment in large
  language models.
\newblock \emph{arXiv preprint arXiv:2609.02526}, 2026.

\bibitem[Gao et~al.(2025)Gao, Lee, Burtch, and Fazelpour]{gao2025caution}
Yuan Gao, Dokyun Lee, Gordon Burtch, and Sina Fazelpour.
\newblock Take caution in using {LLMs} as human surrogates.
\newblock \emph{Proceedings of the National Academy of Sciences}, 122\penalty0
  (24):\penalty0 e2501660122, 2025.

\bibitem[Garz{\'o}n et~al.(2026)Garz{\'o}n, Baron, Grari, Kamphorst, Bernstein,
  and Detyniecki]{garzon2026demographics}
Rub{\'e}n Garz{\'o}n, Pauline Baron, Vincent Grari, Jonne Kamphorst, Michael
  Bernstein, and Marcin Detyniecki.
\newblock From demographics to survey anchors: Evaluating {LLM} agents for
  modeling retirement attitudes.
\newblock \emph{arXiv preprint arXiv:2605.16303}, 2026.

\bibitem[Gui and Toubia(2023)]{gui2025challenge}
George Gui and Olivier Toubia.
\newblock The challenge of using {LLMs} to simulate human behavior: A causal
  inference perspective.
\newblock \emph{arXiv preprint arXiv:2312.15524}, 2023.

\bibitem[Gupta et~al.(2024)Gupta, Shrivastava, Deshpande, Kalyan, Clark,
  Sabharwal, and Khot]{gupta2024bias}
Shashank Gupta, Vaishnavi Shrivastava, Ameet Deshpande, Ashwin Kalyan, Peter
  Clark, Ashish Sabharwal, and Tushar Khot.
\newblock Bias runs deep: Implicit reasoning biases in persona-assigned {LLMs}.
\newblock In \emph{International Conference on Learning Representations}, 2024.

\bibitem[Hartley et~al.(2025)Hartley, Hamill, Seddon, Batra, Okhrati, and
  Khraishi]{hartley2025personality}
John Hartley, Conor Hamill, Dale Seddon, Devesh Batra, Ramin Okhrati, and Raad
  Khraishi.
\newblock How personality traits shape {LLM} risk-taking behaviour.
\newblock In \emph{Findings of the Association for Computational Linguistics:
  ACL 2025}, pages 21068--21092, 2025.

\bibitem[Hu and Collier(2024)]{hu2024personaeffect}
Tiancheng Hu and Nigel Collier.
\newblock Quantifying the persona effect in {LLM} simulations.
\newblock In \emph{Proceedings of the 62nd Annual Meeting of the Association
  for Computational Linguistics (Volume 1: Long Papers)}, pages 10289--10307,
  2024.

\bibitem[Hu et~al.(2026)Hu, Baumann, Lupo, Collier, Hovy, and
  R{\"o}ttger]{hu2026simbench}
Tiancheng Hu, Joachim Baumann, Lorenzo Lupo, Nigel Collier, Dirk Hovy, and Paul
  R{\"o}ttger.
\newblock {SimBench}: Benchmarking the ability of large language models to
  simulate human behaviors.
\newblock In \emph{International Conference on Learning Representations}, 2026.

\bibitem[Iwamoto et~al.(2026)Iwamoto, Ishihara, and Ida]{iwamoto2026risk}
Ryota Iwamoto, Takunori Ishihara, and Takanori Ida.
\newblock Comparing risk preferences and reference dependence in humans and
  {AI}: A persona-based approach with fine-tuning.
\newblock \emph{Journal of Behavioral and Experimental Economics},
  124:\penalty0 102634, 2026.

\bibitem[Jia et~al.(2026)Jia, Chen, Sharma, and
  Diaz-Rodriguez]{jia2026personas}
Mumin Jia, Yilin Chen, Divya Sharma, and Jairo Diaz-Rodriguez.
\newblock When can digital personas reliably approximate human survey findings?
\newblock \emph{arXiv preprint arXiv:2605.10659}, 2026.

\bibitem[Kinzinger and Hartmann(2026)]{kinzinger2026synthetic}
Leonard Kinzinger and Jochen Hartmann.
\newblock Synthetic personalities: How well can {LLMs} mimic individual
  respondents using socio-economic microdata?
\newblock \emph{arXiv preprint arXiv:2606.04592}, 2026.

\bibitem[Korinek(2023)]{korinek2023generative}
Anton Korinek.
\newblock Generative {AI} for economic research: Use cases and implications for
  economists.
\newblock \emph{Journal of Economic Literature}, 61\penalty0 (4):\penalty0
  1281--1317, 2023.

\bibitem[Li et~al.(2025)Li, Chen, Namkoong, and Peng]{li2025promise}
Ang Li, Haozhe Chen, Hongseok Namkoong, and Tianyi Peng.
\newblock {LLM} generated persona is a promise with a catch.
\newblock In \emph{Advances in Neural Information Processing Systems},
  volume~38, pages 73875--73896, 2025.

\bibitem[Liu et~al.(2025{\natexlab{a}})Liu, Tang, Yang, and
  Tam]{liu2025riskpersona}
Jiaxin Liu, Yixuan Tang, Yi~Yang, and Kar~Yan Tam.
\newblock Evaluating and aligning human economic risk preferences in {LLMs}.
\newblock In \emph{Proceedings of the 2025 Conference on Empirical Methods in
  Natural Language Processing}, pages 18185--18199, 2025{\natexlab{a}}.

\bibitem[Liu et~al.(2025{\natexlab{b}})Liu, Geng, Peterson, Sucholutsky, and
  Griffiths]{liu2025rational}
Ryan Liu, Jiayi Geng, Joshua~C. Peterson, Ilia Sucholutsky, and Thomas~L.
  Griffiths.
\newblock Large language models assume people are more rational than we really
  are.
\newblock In \emph{International Conference on Learning Representations},
  2025{\natexlab{b}}.

\bibitem[Lopez-Lira et~al.(2025)Lopez-Lira, Tang, and
  Zhu]{lopezlira2025memorization}
Alejandro Lopez-Lira, Yuehua Tang, and Mingyin Zhu.
\newblock The memorization problem: Can we trust {LLMs}' economic forecasts?
\newblock \emph{arXiv preprint arXiv:2504.14765}, 2025.

\bibitem[Ludwig et~al.(2026)Ludwig, Mullainathan, and
  Rambachan]{ludwig2026econometric}
Jens Ludwig, Sendhil Mullainathan, and Ashesh Rambachan.
\newblock Large language models: An applied econometric framework.
\newblock \emph{Annual Review of Economics}, 18\penalty0 (1):\penalty0
  283--316, 2026.

\bibitem[Lukauskas and {\v{S}}arkauskait{\.e}(2026)]{lukauskas2026plausible}
Mantas Lukauskas and Viktorija {\v{S}}arkauskait{\.e}.
\newblock Plausible but not valid: A psychometric audit of {LLMs} as synthetic
  survey respondents.
\newblock \emph{arXiv preprint arXiv:2608.14606}, 2026.

\bibitem[Manning and Horton(2026)]{manning2026gsa}
Benjamin~S. Manning and John~J. Horton.
\newblock General social agents.
\newblock Working Paper 34937, National Bureau of Economic Research, 2026.

\bibitem[Manning et~al.(2024)Manning, Zhu, and Horton]{manning2024automated}
Benjamin~S. Manning, Kehang Zhu, and John~J. Horton.
\newblock Automated social science: Language models as scientist and subjects.
\newblock Working Paper 32381, National Bureau of Economic Research, 2024.

\bibitem[Mantel(1967)]{mantel1967detection}
Nathan Mantel.
\newblock The detection of disease clustering and a generalized regression
  approach.
\newblock \emph{Cancer Research}, 27\penalty0 (2):\penalty0 209--220, 1967.

\bibitem[Mei et~al.(2024)Mei, Xie, Yuan, and Jackson]{mei2024turing}
Qiaozhu Mei, Yutong Xie, Walter Yuan, and Matthew~O. Jackson.
\newblock A {Turing} test of whether {AI} chatbots are behaviorally similar to
  humans.
\newblock \emph{Proceedings of the National Academy of Sciences}, 121\penalty0
  (9):\penalty0 e2313925121, 2024.

\bibitem[Mili{\v{c}}ka et~al.(2024)Mili{\v{c}}ka, Marklov{\'a}, VanSlambrouck,
  Posp{\'\i}{\v{s}}ilov{\'a}, {\v{S}}imsov{\'a}, Harvan, and
  Drobil]{milicka2024downplay}
Ji{\v{r}}{\'\i} Mili{\v{c}}ka, Anna Marklov{\'a}, Kl{\'a}ra VanSlambrouck, Eva
  Posp{\'\i}{\v{s}}ilov{\'a}, Jana {\v{S}}imsov{\'a}, Samuel Harvan, and
  Ond{\v{r}}ej Drobil.
\newblock Large language models are able to downplay their cognitive abilities
  to fit the persona they simulate.
\newblock \emph{PLOS ONE}, 19\penalty0 (3):\penalty0 e0298522, 2024.

\bibitem[Park et~al.(2024{\natexlab{a}})Park, Zou, Kamphorst, Egan, Shaw, Hill,
  Cai, Morris, Liang, Willer, and Bernstein]{park2024generative}
Joon~Sung Park, Carolyn~Q. Zou, Jonne Kamphorst, Niles Egan, Aaron Shaw,
  Benjamin~Mako Hill, Carrie Cai, Meredith~Ringel Morris, Percy Liang, Robb
  Willer, and Michael~S. Bernstein.
\newblock {LLM} agents grounded in self-reports enable general-purpose
  simulation of individuals.
\newblock \emph{arXiv preprint arXiv:2411.10109}, 2024{\natexlab{a}}.

\bibitem[Park et~al.(2024{\natexlab{b}})Park, Schoenegger, and
  Zhu]{park2024diminished}
Peter~S. Park, Philipp Schoenegger, and Chongyang Zhu.
\newblock Diminished diversity-of-thought in a standard large language model.
\newblock \emph{Behavior Research Methods}, 56\penalty0 (6):\penalty0
  5754--5770, 2024{\natexlab{b}}.

\bibitem[Peng et~al.(2026)Peng, Brucks, Gui, Merlau, Fan, Ben~Sliman, Johnson,
  Althenayyan, Bellezza, Donati, Fong, Friedman, Guevara, Hussein, Jerath,
  Kogut, Kumar, Lane, Li, Morwitz, Netzer, Perkowski, and
  Toubia]{peng2026funhouse}
Tianyi Peng, Melanie Brucks, George Gui, Daniel~J. Merlau, Grace~Jiarui Fan,
  Malek Ben~Sliman, Eric~J. Johnson, Abdullah Althenayyan, Silvia Bellezza,
  Dante Donati, Hortense Fong, Elizabeth Friedman, Ariana Guevara, Mohamed
  Hussein, Kinshuk Jerath, Bruce Kogut, Akshit Kumar, Kristen Lane, Hannah Li,
  Vicki Morwitz, Oded Netzer, Patryk Perkowski, and Olivier Toubia.
\newblock Digital twins are funhouse mirrors: Five systematic distortions.
\newblock \emph{Science Advances}, 12\penalty0 (36):\penalty0 eaeh8260, 2026.

\bibitem[Rammstedt and John(2007)]{rammstedt2007bfi10}
Beatrice Rammstedt and Oliver~P. John.
\newblock Measuring personality in one minute or less: A 10-item short version
  of the {B}ig {F}ive {I}nventory in {E}nglish and {G}erman.
\newblock \emph{Journal of Research in Personality}, 41\penalty0 (1):\penalty0
  203--212, 2007.

\bibitem[Ross et~al.(2024)Ross, Kim, and Lo]{ross2024economicus}
Jillian Ross, Yoon Kim, and Andrew~W. Lo.
\newblock {LLM} economicus? {M}apping the behavioral biases of {LLMs} via
  utility theory.
\newblock In \emph{Conference on Language Modeling}, 2024.

\bibitem[Santurkar et~al.(2023)Santurkar, Durmus, Ladhak, Lee, Liang, and
  Hashimoto]{santurkar2023whose}
Shibani Santurkar, Esin Durmus, Faisal Ladhak, Cinoo Lee, Percy Liang, and
  Tatsunori Hashimoto.
\newblock Whose opinions do language models reflect?
\newblock In \emph{Proceedings of the 40th International Conference on Machine
  Learning}, volume 202 of \emph{Proceedings of Machine Learning Research},
  pages 29971--30004, 2023.

\bibitem[Sen et~al.(2025)Sen, Lutz, Rogers, Garcia, and
  Strohmaier]{sen2025margins}
Indira Sen, Marlene Lutz, Elisa Rogers, David Garcia, and Markus Strohmaier.
\newblock Missing the margins: A systematic literature review on the
  demographic representativeness of {LLMs}.
\newblock In \emph{Findings of the Association for Computational Linguistics:
  ACL 2025}, pages 24263--24289, 2025.

\bibitem[Shapley(1953)]{shapley1953value}
Lloyd~S. Shapley.
\newblock A value for $n$-person games.
\newblock In Harold~W. Kuhn and Albert~W. Tucker, editors, \emph{Contributions
  to the Theory of Games II}, volume~28 of \emph{Annals of Mathematics
  Studies}, pages 307--317. Princeton University Press, 1953.

\bibitem[Shi and Haupt(2026)]{shi2026collapse}
Yuanming Shi and Andreas Haupt.
\newblock The collapse of heterogeneity in {Silicon Philosophers}.
\newblock In \emph{Proceedings of the 2026 ACM Conference on Fairness,
  Accountability, and Transparency}, pages 6280--6304, 2026.

\bibitem[Stango and Zinman(2023)]{stango2023behavioral}
Victor Stango and Jonathan Zinman.
\newblock We are all behavioural, more, or less: A taxonomy of consumer
  decision-making.
\newblock \emph{The Review of Economic Studies}, 90\penalty0 (3):\penalty0
  1470--1498, 2023.

\bibitem[Taday~Morocho et~al.(2026)Taday~Morocho, Cima, Fagni, Avvenuti, and
  Cresci]{morocho2026assessing}
Erika~Elizabeth Taday~Morocho, Lorenzo Cima, Tiziano Fagni, Marco Avvenuti, and
  Stefano Cresci.
\newblock Assessing the reliability of persona-conditioned {LLMs} as synthetic
  survey respondents.
\newblock In \emph{Companion Proceedings of the ACM Web Conference 2026}, pages
  320--329, 2026.

\bibitem[Toubia et~al.(2025)Toubia, Gui, Peng, Merlau, Li, and
  Chen]{toubia2025twin}
Olivier Toubia, George Gui, Tianyi Peng, Daniel~J. Merlau, Ang Li, and Haozhe
  Chen.
\newblock Database report: {Twin-2K-500}: A data set for building digital twins
  of over 2,000 people based on their answers to over 500 questions.
\newblock \emph{Marketing Science}, 44\penalty0 (6):\penalty0 1446--1455, 2025.

\bibitem[Wang et~al.(2025)Wang, Morgenstern, and Dickerson]{wang2025misportray}
Angelina Wang, Jamie Morgenstern, and John~P. Dickerson.
\newblock Large language models that replace human participants can harmfully
  misportray and flatten identity groups.
\newblock \emph{Nature Machine Intelligence}, 7\penalty0 (3):\penalty0
  400--411, 2025.

\bibitem[Wang et~al.(2026)Wang, Hunt, Tang, and
  Joseph]{wang2026substitutability}
Steven Wang, Kyle Hunt, Shaojie Tang, and Kenneth Joseph.
\newblock When can {LLM} digital twins reduce human measurement? {F}rom
  behavioral fidelity to statistical substitutability.
\newblock \emph{arXiv preprint arXiv:2609.07987}, 2026.

\bibitem[Xiao et~al.(2026)Xiao, Zhang, Yang, Ma, Xuan, and
  Huang]{xiao2026chameleon}
Yunze Xiao, Vivienne~J. Zhang, Chenghao Yang, Ningshan Ma, Weihao Xuan, and
  Jen-tse Huang.
\newblock The {Chameleon's} limit: Investigating persona collapse and
  homogenization in large language models.
\newblock \emph{arXiv preprint arXiv:2604.24698}, 2026.

\bibitem[Yoshikawa et~al.(2026)Yoshikawa, Takayama, and
  Yamazaki]{yoshikawa2026persona}
Katsumasa Yoshikawa, Junya Takayama, and Takato Yamazaki.
\newblock Persona-aware evaluation of cognitive bias in {LLMs}: From benchmark
  to applied decision-making.
\newblock In \emph{Proceedings of the Fifteenth Language Resources and
  Evaluation Conference}, pages 4213--4225, 2026.

\bibitem[Youden(1950)]{youden1950index}
William~J. Youden.
\newblock Index for rating diagnostic tests.
\newblock \emph{Cancer}, 3\penalty0 (1):\penalty0 32--35, 1950.

\bibitem[Zhu et~al.(2025)Zhu, Yan, and Griffiths]{yan2025arithmetic}
Jian-Qiao Zhu, Haijiang Yan, and Thomas~L. Griffiths.
\newblock Language models trained to do arithmetic predict human risky and
  intertemporal choice.
\newblock In \emph{International Conference on Learning Representations}, 2025.

\end{thebibliography}

\appendix
\raggedbottom

\clearpage
\section{Additional related work}
\label{app:related}

\textbf{Simulating people with LLMs.} In large-scale replications, LLMs reproduce many scenario-based
experimental effects \citep{cui2025replication}, yet demographic prompts give the model no reliable basis for
within-sample heterogeneity \citep{manning2026gsa}. Several studies document how persona-conditioned responses narrow.
Answers show strong question-specific bias with low variance \citep{boelaert2025machinebias}, judgments become
over-correlated across domains \citep{shi2026collapse}, and what variation remains tracks coarse demographic
stereotypes instead of individual differences \citep{xiao2026chameleon}. Persona prompting does not reliably improve alignment with survey
distributions \citep{morocho2026assessing}, and whether persona attributes help depends on how much
humans vary on a question \citep{frohling2026persona}.

\textbf{Behavioral biases in LLMs.} Earlier studies test LLMs on economic rationality, strategic
behavior, and behavioral biases without matching answers to individual respondents
\citep{chen2023rationality,mei2024turing,ross2024economicus}, or show that assigned personality traits
shift LLM risk-taking \citep{hartley2025personality}; \citet{binz2025centaur} fine-tune a model to predict
human choices in psychology experiments. Here, each verdict is scored against the same respondent's
answers.

\textbf{Representation and calibration.} Whether LLM outputs represent demographic groups evenly is a
known concern \citep{santurkar2023whose,sen2025margins}: a model can match national aggregates while
misrepresenting subgroups \citep{bollen2026local}, and LLM risk scores for individuals can carry
predictive signal yet be widely miscalibrated \citep{cruz2024risk}.

\section{The 14 biases}
\label{app:biases}

Table~\ref{tab:biases} gives the instrument and scoring rule for each outcome. All binary verdicts follow one sign
convention, with 1 marking the biased response, and we score human and model answers with the same code, using the
source study's rules \citep{stango2023behavioral}.

Two sample sizes appear in the table. The first, $n$, counts the valid wave-2 responses used for prevalence. The
second, $n_{\text{pair}}$, keeps only respondents who also have a valid wave-1 response and a valid prediction from
every condition; this common retest sample is the one behind the test--retest benchmark. The gap between the two
is largest for certainty preference, because its scoring rule excludes respondents who switch more than once in
the elicitation lists.

Because we drop three of the source study's 17 outcomes (Section~\ref{sec:setup}), we recompute every count over
the 14 text-administered outcomes.

\begin{table}[H]
\caption{\textbf{The 14 biases, scoring rules, and coverage.} Rate is the human wave-2 prevalence.
EGB is exponential-growth bias and LLN the law of large numbers.}
\label{tab:biases}
\centering\footnotesize\setlength{\tabcolsep}{3.5pt}
\begin{tabular}{@{}L{3.05cm}L{3.15cm}L{3.15cm}rrr@{}}
\toprule
Bias & Instrument & Biased if & Rate & $n$ & $n_{\text{pair}}$\\
\midrule
\multicolumn{6}{@{}l}{\textit{Present bias}}\\
Present bias --- money & 24 convex time-budget allocations & less patient when the earlier date is today & 0.248 & 819 & 782\\
Present bias --- consumption & two snack choices, now vs later & unhealthy now \emph{and} healthy later & 0.168 & 829 & 821\\
\addlinespace[2pt]
\multicolumn{6}{@{}l}{\textit{Inconsistent choice}}\\
Narrow bracketing & Rabin--Weizs\"acker pairs, shown jointly & dominated pair, or not the both-safe pair & 0.624 & 837 & 820\\
\addlinespace[2pt]
\multicolumn{6}{@{}l}{\textit{Risk}}\\
Loss aversion & Fehr--Goette single vs six \$80/\$50 gambles & rejects one or both & 0.602 & 845 & 843\\
Certainty preference & two probability-equivalent lists & Callen certainty premium $>0$ & 0.762 & 629 & 526\\
Ambiguity aversion & Ellsberg two-bag choice & chooses the known bag & 0.764 & 821 & 818\\
\addlinespace[2pt]
\multicolumn{6}{@{}l}{\textit{Overconfidence}}\\
Overconfidence --- level & predicted vs actual numeracy score & predicted $>$ actual & 0.339 & 813 & 799\\
Overconfidence --- precision & two subjective probability distributions & places 100\% on a single bin & 0.454 & 775 & 748\\
Overconfidence --- relative & self-ranked percentile before the test & self-rank $>$ realized rank & 0.730 & 818 & 817\\
\addlinespace[2pt]
\multicolumn{6}{@{}l}{\textit{Math and statistics}}\\
Non-belief in the LLN & 1,000 coin flips, central interval & $\Pr(481\text{--}519) < 78\%$ & 0.852 & 819 & 809\\
Gambler's fallacy & next flip after a run & answer $<50$ & 0.240 & 817 & 815\\
EGB --- loans & monthly payment on a stated loan & implied APR $>$ stated APR & 0.527 & 783 & 736\\
EGB --- assets & \$200 at 10\% compounded for two years & answer $<242$ & 0.414 & 735 & 688\\
\addlinespace[2pt]
\multicolumn{6}{@{}l}{\textit{Attention}}\\
Limited attention & four regret/inattention items & any flagged response & 0.444 & 829 & 819\\
\bottomrule
\end{tabular}
\end{table}

\section{Prompt construction and administration}
\label{app:prompts}

The prompt text below is verbatim; persona values in the examples are invented.

\textbf{Prompt.} Each prompt had a system section, holding an instruction preamble and the respondent
profile, and a user section, holding one survey item and its response format. The task statement and
the empty profile of the no-identity condition follow \citet{peng2026funhouse}; the section structure
and guidelines are taken from, and three of the instructions adapted from, \citet{li2025promise}. Because prompt order, labels, and framing can themselves bias LLM answers
\citep{brucks2025prompt}, items are worded and formatted identically in every condition. The preamble was identical in A1--A4:

\begin{scriptsize}
\begin{verbatim}
### TASK ###
You are simulating one real survey respondent. The 'Respondent Profile' below describes a
participant in a study of household financial decision-making; it may contain their
background, their personality and ability scores, and their answers to an earlier set of
questions. Answer the 'New Survey Question' exactly as that person would answer it.

### GUIDELINES ###
1. Be Faithful to the Persona: Ensure your answer is consistent with the persona's data.
2. Focus on Relevant Aspects: Center your reasoning on the relevant factors that would
   influence the persona's opinion on that topic.
3. Be Objective: Avoid injecting personal bias or overly politically correct views that
   may not align with the persona's standpoint.

### INSTRUCTIONS ###
- Give exactly one answer per question, in the format that question asks for.
- If more than one answer is plausible for this person, select among them in proportion to
  how likely each is. Do not default to the most common or the most sensible answer.
- Always answer, even in unclear cases - treat it as a forced-choice survey.
- Reason internally. Do not include your reasoning in the response.

# Respondent Profile
\end{verbatim}
\end{scriptsize}

A5 adds two sentences after the task paragraph: \emph{``The `Respondent Profile' includes this
person's responses to an earlier set of questions. Use them as evidence of how the persona answers
questions of this kind.''}

\textbf{Respondent profile.} In A1 the profile reads \texttt{[No respondent profile provided.]} A2
adds the 16 wave-2 demographic fields, shown as 13 labeled lines because three of the fields are
numeric codes behind the education and income labels. A missing field is omitted rather than imputed.

\begin{scriptsize}
\begin{verbatim}
## Demographics
- Age: 54
- Gender: Male
- Race: White
- Hispanic or Latino origin: No
- Highest level of education completed: Some college, no degree
- Total household income last year, before taxes: $40,000 to $49,999
- Employment status: Working now
- Work arrangement: Works for someone else
- Occupational group: Office and Administrative Support
- Marital status: Married or living with a partner
- Housing: Owns
- Number of other people living in the household: 2
- Born in the United States: Yes
\end{verbatim}
\end{scriptsize}

A3 adds the BFI-10 personality block \citep{rammstedt2007bfi10}; missing traits are handled the same
way.

\begin{scriptsize}
\begin{verbatim}
## Personality

This person completed the BFI-10, a standard ten-item measure of the five major personality
traits. Each trait is measured by two statements - one worded positively, one worded
negatively - which the person rated on a five-point scale from "Disagree strongly" to
"Agree strongly". The two ratings are combined, reversing the negatively worded one, into a
single score from 1.0 to 5.0. A score of 3.0 is neutral; higher means the trait is more
strongly present, lower means less.

- Openness - curiosity, imagination, receptiveness to new experiences and ideas.
  Rated on: "has an active imagination" and "has few artistic interests".
  Score: 3.5 out of 5

  [the remaining four traits follow in the same form]
\end{verbatim}
\end{scriptsize}

A4 adds the four cognitive scores, each reported against its scale.

\begin{scriptsize}
\begin{verbatim}
## Measured cognitive skills

This person took four short tests as part of the same survey. Their results:

- Numeracy - two questions about division and percentages.
  Score: 2 out of 2 correct.

- Financial literacy - three standard questions on compound interest,
  inflation, and investment diversification.
  Score: 3 out of 3 correct.

- Fluid intelligence - a number-series test in which the respondent
  supplies the next number in a sequence.
  Score: 10 out of 15 correct.

- Executive function - a timed test requiring the respondent to name the
  ink colour of a colour word while ignoring the word itself, as many as
  possible in two minutes.
  Score: 61 correct in two minutes. The average for this survey's
  respondents was 70.
\end{verbatim}
\end{scriptsize}

LLMs can answer below their own ability to fit a simulated persona \citep{milicka2024downplay};
Figure~\ref{fig:incap} tests whether stated scores have this effect here. The profile's numeracy score
counts the two items on division and percentages; Figure~\ref{fig:incap}(b) also counts the
compound-interest item shown in panel~(a).

A5 adds the respondent's wave-1 answers, excluding the withheld items described below, in this form:

\begin{scriptsize}
\begin{verbatim}
## How this person has answered questions before - use as evidence

These are this person's own responses, recorded on an earlier occasion. They show how this
person reads questions, what they tend to prefer, and how carefully they work things out.

Q7. Suppose you had $100 in a savings account and the interest rate was 2% per year. After
5 years, how much do you think you would have in the account if you left the money to grow?
  1. More than $102
  2. Exactly $102
  3. Less than $102
  4. Don't know
Answer: 1 (More than $102)
\end{verbatim}
\end{scriptsize}

\textbf{Survey item and response format.} Each user section carried one item and a response format:

\begin{scriptsize}
\begin{verbatim}
# New Survey Question

Q85. Imagine that we had a computer 'flip' a fair coin: a fair coin is one with a 50-50
chance of coming up heads or tails. Say that the computer flips the coin 10 times. The first
9 are all heads. What are the chances, in percentage terms, that the 10th flip will be a
head?

Format Instructions:
Respond with only a JSON object in exactly this form, and nothing else:
{"Q85": <answer>}
\end{verbatim}
\end{scriptsize}

Item wording follows the source instrument \citep{stango2023behavioral}, adapted for text
administration; the income-change distribution (Q28b) asks about realized rather than expected change.
Each Rabin--Weizs\"acker task is shown on a single screen with an instruction to consider both
decisions together. Items that refer to earlier answers (Q27, Q28, Q30 and Q88) restate those answers
from the same condition, and conditional follow-up items are asked only when triggered.

\textbf{Withholding rules.} The prediction target never enters a prompt: wave-2 answers and all
derived bias variables are excluded from every condition, and profiles contain structured fields only,
with no names or free text. In A4 and A5, a cognitive score is hidden on the screens that would reveal
its answer: numeracy on Q4--Q6, Q27, Q28 and Q28b; financial literacy on Q7--Q9; and fluid
intelligence on the number series, Q29 and Q30. The A5 history omits the entire item block used to
score the target bias.

\textbf{Administration.} Each item was answered in a separate, single-turn call to a Claude Code
sub-agent running Claude Opus 4.8 (\texttt{claude-opus-4-8}) in August 2026, with a fresh context and
no conversation history, except where items must be answered together (\S\ref{sec:setup}). The system and user sections were sent together, followed by the closing
instruction:

\begin{scriptsize}
\begin{verbatim}
(Answer as this persona. You are a plain-text respondent: use no tools, reason internally,
and output ONLY the JSON object requested, nothing else.)
\end{verbatim}
\end{scriptsize}

Claude Opus 4.8 does not accept sampling parameters, so none were set. One reply was generated for
each respondent, item, and condition, and the answer was read from the JSON object in the reply.

\section{Estimators}
\label{app:estimators}

We define each estimator explicitly, because an individual-level score is easy to interpret only on a scale on
which a predictor that ignores the person scores zero.

For a bias $b$ with human wave-2 verdict $Y\in\{0,1\}$ and a predictor $\hat Y$:

\begin{align}
\text{accuracy} &= \Pr(\hat Y = Y),\\
\text{TPR} = \Pr(\hat Y {=} 1 \mid Y {=} 1), \qquad
\text{TNR} &= \Pr(\hat Y {=} 0 \mid Y {=} 0),\\
J &= \text{TPR} + \text{TNR} - 1.
\end{align}

Informedness, also known as Youden's index \citep{youden1950index}, has a null value of $0$. Raw accuracy has no
single null value. A majority-class predictor scores $\max(p_b,1-p_b)$ and a random draw at the population rate
scores $p_b^2+(1-p_b)^2$. Both scores depend only on the prevalence $p_b$.

We express a condition's value $v$ as a share of the human benchmark,
$(v-\text{null})/(\text{benchmark}-\text{null})$. The subtraction is essential. Without it, a predictor at the null would still receive a large share of the human benchmark. For this reason Table~\ref{tab:accuracy} reports raw accuracy descriptively and does not normalize it as a share of the benchmark; the comparisons rest on the chance-corrected
rows.

\section{Samples}
\label{app:samples}

No single sample suits every result, because different analyses need different observed cells. The text, a
caption, or this appendix states the sample behind each analysis.

\textbf{Available case} ($n=629$--845 per bias). Everyone with a valid wave-2 answer for the bias is in this sample. We use it for prevalence and the other
marginal distributions, which need no wave-1 response.

\textbf{Common retest} ($n=526$--843 per bias). Here we also require a valid wave-1 answer and a valid prediction from every condition. The primary accuracy and
informedness comparisons use this sample, because it defines the human benchmark and all model conditions on the
same people.

\textbf{Counts} ($n=696$). For the bias-count distribution (Figure~\ref{fig:bcount}) we keep respondents with all 14 outcomes available in
every condition. Each count is then a whole number out of 14, and every condition is compared on the same people.
In this sample, 3.0\% of human wave-2 cells are filled from the same person's wave-1 answer, and the human mean is
7.08 (SD 1.90). The source study instead multiplies each person's share of valid outcomes by 14. Applying this rule to all 845
respondents gives the same pattern, with human 7.15 (SD 1.91), A1 SD 0.73, A2--A4 SD 1.53--1.62,
and A5 7.32 (SD 1.93). Group prevalence (panel~(b) of Figure~\ref{fig:columns}) uses this all-respondent
version.

\textbf{Primary sample} ($n=829$). For the primary classifier and residual analyses, a respondent needs at least 7 of the 14 biases to be valid in
both human waves and in every model condition, and is then compared on their common observed cells. A stricter
requirement of 8 biases ($n=818$) or 10 biases ($n=794$) changes no headline estimate by more than 0.011.
Identification has its own validity rule, which retains 828 profiles, or 788 in the demographic-matched
version.

\textbf{Complete-profile sensitivity checks.} If all 14 binary outcomes must be present, 389 profiles remain for the residual and classifier analyses and 386 for
identification. Without certainty preference, 532 and 531 profiles remain, respectively.

\textbf{Balanced} ($n=518$). The variance decomposition needs all 14 outcomes to be valid for the human target and for every model condition.
With this complete rectangle, no missingness has to be assigned to a person or a bias component. Its human mean is
6.92. Means from different samples answer different questions, and we do not combine them.

\textbf{Auxiliary samples.} Four analyses use a validity rule of their own. The block-Shapley decomposition in Table~\ref{tab:blockcoef}
needs complete \emph{predictors}, not complete outcomes. Requiring all 18 attributes keeps 757 respondents. The 18 attributes are nine demographic variables (age, gender, White, Hispanic origin, education, income, employment, marital
status, and household size), the five traits, and the four cognitive scores. The other 88 respondents lack at
least one attribute, almost always a Big Five trait or a cognitive score.
The respondent-level comparison (Table~\ref{tab:person-gains}) retains 837 respondents for binary
accuracy and 821 for graded error, which excludes present bias for consumption (no graded score) and certainty
preference. Group accuracy in panel~(a) of Figure~\ref{fig:columns} requires at least 8 valid biases
and retains 818 respondents.
Group informedness (Table~\ref{tab:groupinf} and Figure~\ref{fig:ladder}) instead pools each bias's
own common-retest sample and so rests on the 845 respondents scorable for at least one bias.

Table~\ref{tab:groupinf} and Table~\ref{tab:grid} leave out Asian (22) and American Indian (12) respondents, whose
group-by-bias cells all have fewer than 40 valid observations. Figure~\ref{fig:columns} instead
combines race and Hispanic origin into four groups:
White non-Hispanic (592), Black non-Hispanic (92), Hispanic of any race (111), and other non-Hispanic
(50, including Asian and American Indian respondents). Income groups are below \$40,000 (328),
\$40,000--\$99,999 (318), and \$100,000 or more (198; one respondent has no income recorded), age groups are
terciles, and education groups are high school or less, some college (including associate degrees),
bachelor's, and postgraduate.

\section{Coverage limits and sensitivity checks}
\label{app:limits}

\textbf{Why we report single attributes and not intersectional groups.} \S\ref{sec:whom} compares groups one
attribute at a time, because intersectional cells have uneven coverage. Requiring at least five personas in each income $\times$ ethnicity $\times$ age $\times$
education $\times$ gender cell keeps 95\% of White personas and 45\% of Black personas, but none of the Asian,
American Indian, or Other personas, and no Black-male cell remains. The extreme cells are also
unstable. Among the 60 cells with at least five people, the split-half reliability of the cell ordering is only
0.26--0.32, so the largest and smallest estimates largely reflect noise; single-attribute gradients avoid both
problems.

\textbf{One mechanically confounded pair is excluded from the Mantel test.}
Overconfidence-level is predicted minus actual numeracy, and the actual score contains the same item that measures
exponential-growth bias on assets. The pair is therefore correlated by construction and is the largest entry in both the human and the synthetic correlation matrices. When the pair is included, four of five conditions
appear to match the human correlation structure at $p<0.05$, and when it is excluded, none do
(Table~\ref{tab:mantel}). \S\ref{sec:population} reports the result without the pair.

\begin{table}[H]
\centering\footnotesize
\caption{\textbf{Mantel test of the correlation structure among biases.} $r$ is the correlation between
the human and synthetic between-bias Spearman correlation matrices, computed on the source study's percentile (tile) bias scores; for present bias on consumption, which has no graded
score, the tile of its binary verdict is used. The two-sided $p$ comes from 5,000 permutations of the bias
labels. Biases whose synthetic scores do not vary have undefined
correlations, so A1 and A2 use fewer pairs.}
\label{tab:mantel}
\begin{tabular}{@{}lrrrrrr@{}}
\toprule
& \multicolumn{3}{c}{All pairs} & \multicolumn{3}{c}{Without the confounded pair}\\
\cmidrule(lr){2-4}\cmidrule(l){5-7}
Condition & Pairs & $r$ & $p$ & Pairs & $r$ & $p$\\
\midrule
A1 & 55 & 0.623 & 0.004 & 54 & 0.209 & 0.138\\
A2 & 78 & 0.176 & 0.288 & 77 & $-0.129$ & 0.462\\
A3 & 91 & 0.363 & 0.016 & 90 & $-0.033$ & 0.820\\
A4 & 91 & 0.278 & 0.046 & 90 & 0.010 & 0.950\\
A5 & 91 & 0.429 & 0.006 & 90 & 0.244 & 0.118\\
\bottomrule
\end{tabular}
\end{table}

\section{Additional tables}
\label{app:extra}

The tables below give the values behind Figures~\ref{fig:prevladder}, \ref{fig:columns}, and \ref{fig:incap}
(Tables~\ref{tab:prev-values}--\ref{tab:incap-values}) and the profile-level checks cited in \S\ref{sec:population}
and \S\ref{sec:ladder}. Table~\ref{tab:complete-profile} gives the
classifier and residual-correlation results on the primary sample and on complete 14-bias profiles
(389 respondents for classifier and residual analyses, 386 for identification), and
Table~\ref{tab:primary-identification} gives identification on the primary sample. The complete-profile results are similar. There A5 recovers 20\% of the human residual correlation, against 24\% in
the primary sample, its classifier AUC is 0.976, and its all-pool identification AUC changes from 0.552 to 0.531.
Certainty preference leaves the fewest valid answers because of its multiple-switcher rule. Without it, the 13-bias
complete sample has 532 respondents, and on that sample A5 recovers 23\% of the human residual correlation, with a
classifier AUC of 0.970 and an identification AUC of 0.544.

\textbf{Reassignment test.} We group respondents into strata by gender, age tercile, two income groups (below and above \$40,000), and three
education bands (below high school, high school to associate degree, bachelor's or higher), and we drop strata with
fewer than six respondents, which leaves 516--823 respondents per bias. Within each stratum we permute A5 verdicts
among respondents and recompute informedness. Over 2,000 permutations the mean is 0.026 (95\% interval
0.010--0.042), against 0.078 observed ($p=0.0005$).

\begin{table}[H]
\centering\footnotesize
\caption{\textbf{Prevalence of each bias by condition (values behind Figure~\ref{fig:prevladder}).}
Each bias uses its available cases ($n=629$--845). Bold marks the condition closest to HUMAN.}
\label{tab:prev-values}
\begin{tabular}{@{}lcccccc@{}}
\toprule
Bias & HUMAN & A1 & A2 & A3 & A4 & A5\\
\midrule
Present bias --- money & 0.248 & 0.719 & 0.663 & 0.662 & 0.621 & \textbf{0.587}\\
Present bias --- consumption & 0.168 & 1.000 & 0.900 & 0.802 & 0.853 & \textbf{0.557}\\
Narrow bracketing & 0.624 & 1.000 & 0.980 & 0.996 & 0.927 & \textbf{0.885}\\
Loss aversion & 0.602 & 0.006 & 0.464 & 0.625 & \textbf{0.581} & 0.447\\
Certainty preference & 0.762 & 0.878 & 0.903 & 0.812 & \textbf{0.712} & 0.254\\
Ambiguity aversion & 0.764 & 1.000 & 0.870 & 0.827 & 0.859 & \textbf{0.798}\\
Overconfidence --- level & 0.339 & 0.004 & \textbf{0.264} & 0.260 & 0.079 & 0.456\\
Overconfidence --- precision & 0.454 & 0.000 & 0.000 & 0.003 & 0.000 & \textbf{0.079}\\
Overconfidence --- relative & 0.730 & 0.006 & 0.022 & 0.026 & 0.055 & \textbf{0.061}\\
Non-belief in the LLN & 0.852 & 1.000 & 1.000 & 1.000 & 1.000 & \textbf{0.991}\\
Gambler's fallacy & 0.240 & 0.000 & 0.007 & 0.000 & \textbf{0.017} & 0.002\\
EGB --- loans & 0.527 & 0.849 & 0.918 & 0.867 & 0.884 & \textbf{0.799}\\
EGB --- assets & 0.414 & 0.020 & \textbf{0.426} & 0.278 & 0.141 & 0.555\\
Limited attention & 0.444 & 0.913 & 0.888 & 0.910 & 0.929 & \textbf{0.791}\\
\bottomrule
\end{tabular}
\end{table}

\begin{table}[H]
\centering\footnotesize
\caption{\textbf{Largest gap between groups (values behind Figure~\ref{fig:columns}).} Each cell is the
highest minus the lowest group value within an attribute: (a) accuracy, on respondents with at least
8 valid biases ($n=818$); (b) attributed bias prevalence, averaged over the 14 biases, on all 845
respondents. Ethnicity groups are defined in Appendix~\ref{app:samples}.}
\label{tab:gap-values}
\begin{tabular}{@{}lcccccc@{}}
\toprule
Attribute & HUMAN & A1 & A2 & A3 & A4 & A5\\
\midrule
\multicolumn{7}{@{}l}{\textit{(a) Accuracy}}\\
Age & 0.027 & 0.045 & 0.040 & 0.026 & 0.014 & 0.026\\
Gender & 0.012 & 0.013 & 0.002 & 0.008 & 0.007 & 0.003\\
Ethnicity \& Hispanic origin & 0.026 & 0.034 & 0.057 & 0.047 & 0.059 & 0.043\\
Income & 0.010 & 0.059 & 0.067 & 0.076 & 0.079 & 0.050\\
Education & 0.021 & 0.052 & 0.063 & 0.054 & 0.049 & 0.053\\
\addlinespace[2pt]
\multicolumn{7}{@{}l}{\textit{(b) Attributed bias prevalence}}\\
Age & 0.020 & 0.015 & 0.027 & 0.036 & 0.038 & 0.034\\
Gender & 0.046 & 0.002 & 0.029 & 0.018 & 0.043 & 0.053\\
Ethnicity \& Hispanic origin & 0.051 & 0.015 & 0.057 & 0.043 & 0.078 & 0.123\\
Income & 0.083 & 0.009 & 0.164 & 0.125 & 0.117 & 0.166\\
Education & 0.085 & 0.007 & 0.202 & 0.164 & 0.157 & 0.197\\
\bottomrule
\end{tabular}
\end{table}

\begin{table}[H]
\centering\footnotesize
\caption{\textbf{Compound-interest errors and numeracy by education (values behind
Figure~\ref{fig:incap}).} (a) Share answering the compound-interest item wrongly. (b) Mean number of
the three numeracy questions answered correctly (0--3). $n$ is the number of respondents in each
group.}
\label{tab:incap-values}
\begin{tabular}{@{}lrcccccc@{}}
\toprule
Education & $n$ & HUMAN & A1 & A2 & A3 & A4 & A5\\
\midrule
\multicolumn{8}{@{}l}{\textit{(a) Compound-interest item wrong}}\\
High school or less & 141 & 0.647 & 0.028 & 0.872 & 0.596 & 0.397 & 0.894\\
Some college & 302 & 0.473 & 0.017 & 0.666 & 0.434 & 0.238 & 0.775\\
Bachelor's & 230 & 0.340 & 0.017 & 0.209 & 0.126 & 0.078 & 0.426\\
Postgraduate & 172 & 0.262 & 0.029 & 0.023 & 0.006 & 0.029 & 0.285\\
\addlinespace[2pt]
\multicolumn{8}{@{}l}{\textit{(b) Numeracy score}}\\
High school or less & 141 & 1.56 & 2.97 & 2.13 & 2.40 & 2.60 & 2.10\\
Some college & 302 & 2.11 & 2.98 & 2.33 & 2.57 & 2.76 & 2.23\\
Bachelor's & 230 & 2.43 & 2.98 & 2.79 & 2.87 & 2.92 & 2.57\\
Postgraduate & 172 & 2.61 & 2.97 & 2.98 & 2.99 & 2.97 & 2.72\\
\bottomrule
\end{tabular}
\end{table}

\begin{table}[H]
\centering\footnotesize
\caption{\textbf{Profile-level checks on the primary and complete-profile samples.} Classifier AUC
measures how easily a respondent-grouped classifier separates synthetic from human bias profiles;
0.5 is indistinguishable and closer to 0.5 is better. Residual correlation measures agreement with human
wave 2 after removing person and bias means; higher is better. HUMAN compares wave 1 with wave 2.
Brackets are 95\% respondent-bootstrap intervals. The primary sample has $n=829$ and the
complete-profile sample $n=389$. Dashes mark values not reported.}
\label{tab:complete-profile}
\setlength{\tabcolsep}{3pt}
\begin{tabular}{@{}lcccc@{}}
\toprule
& \multicolumn{2}{c}{Classifier AUC} & \multicolumn{2}{c}{Residual correlation}\\
\cmidrule(lr){2-3}\cmidrule(l){4-5}
Condition & Primary & Complete & Primary & Complete\\
\midrule
HUMAN & 0.476 [0.451, 0.500] & 0.510 [0.476, 0.544] & 0.287 [0.266, 0.308] & 0.297 [0.267, 0.325]\\
A1 & 1.000 [1.000, 1.000] & 1.000 [1.000, 1.000] & $-0.004$ [$-0.026$, 0.017] & ---\\
A2 & 0.996 [0.993, 0.998] & --- & 0.008 [$-0.013$, 0.029] & ---\\
A3 & 0.994 [0.991, 0.996] & --- & 0.022 [0.001, 0.044] & ---\\
A4 & 0.994 [0.991, 0.996] & --- & 0.023 [0.002, 0.043] & 0.040 [0.012, 0.067]\\
A5 & 0.976 [0.970, 0.982] & 0.976 [0.967, 0.984] & 0.070 [0.049, 0.090] & 0.059 [0.029, 0.091]\\
\bottomrule
\end{tabular}
\end{table}
\begin{table}[H]
\centering\footnotesize
\caption{\textbf{Identification on the primary sample.} Identification AUC is
the probability that the correct person's profile is closer than an alternative, with half credit
for tied distances. All-pool comparisons use $n=828$ profiles; demographic-matched comparisons use
$n=788$. Brackets are 95\% respondent-bootstrap intervals. These are the estimates discussed in
\S\ref{sec:ladder}; the complete-profile sensitivity estimate is reported in the text of this appendix.}
\label{tab:primary-identification}
\begin{tabular}{@{}lcc@{}}
\toprule
Condition & All-pool AUC [95\% interval] & Matched AUC [95\% interval]\\
\midrule
HUMAN & 0.718 [0.699, 0.737] & 0.706 [0.685, 0.726]\\
A1 & 0.514 [0.495, 0.534] & 0.513 [0.492, 0.535]\\
A2 & 0.518 [0.498, 0.538] & 0.513 [0.491, 0.534]\\
A3 & 0.524 [0.505, 0.543] & 0.520 [0.499, 0.542]\\
A4 & 0.533 [0.514, 0.553] & 0.530 [0.509, 0.551]\\
A5 & 0.552 [0.534, 0.572] & 0.538 [0.514, 0.560]\\
\bottomrule
\end{tabular}
\end{table}

\section{Continuous scores}
\label{app:continuous}

Thirteen of the 14 biases have a graded score underneath the binary verdict (present bias on
consumption is a pure two-item construct and has none). Table~\ref{tab:continuous} reports, per
bias, how well each side's continuous score discriminates the respondent's wave-2 verdict (AUC,
null 0.5). AUC does not depend on prevalence, so it cannot be inflated the way raw accuracy is.

\textbf{Orientation.} For non-belief in the law of large numbers, gambler's fallacy, and
exponential-growth bias on assets, lower scores indicate more bias. We orient these scores before
computing AUC.

\begin{table}[H]
\caption{\textbf{Agreement on the continuous scores underlying 13 verdicts.} AUC measures how well
each condition's continuous score discriminates the respondent's wave-2 binary verdict; HUMAN uses the
person's wave-1 score. A5 reaches 0.33 of the human benchmark. Per-bias common samples range from $n=526$ to $n=843$.}
\label{tab:continuous}
\centering\footnotesize\setlength{\tabcolsep}{5pt}
\begin{tabular}{@{}lrrrrrr@{}}
\toprule
Bias & HUMAN & A1 & A2 & A3 & A4 & A5\\
\midrule
Present bias --- money & 0.485 & 0.507 & 0.462 & 0.523 & 0.479 & 0.512\\
Loss aversion & 0.694 & 0.502 & 0.459 & 0.505 & 0.464 & 0.548\\
Narrow bracketing & 0.659 & 0.515 & 0.511 & 0.532 & 0.558 & 0.545\\
Certainty preference & 0.564 & 0.505 & 0.463 & 0.434 & 0.541 & 0.521\\
Ambiguity aversion & 0.628 & 0.534 & 0.475 & 0.520 & 0.462 & 0.537\\
Overconfidence --- level & 0.663 & 0.503 & 0.554 & 0.545 & 0.506 & 0.602\\
Overconfidence --- precision & 0.732 & 0.500 & 0.500 & 0.499 & 0.500 & 0.523\\
Overconfidence --- relative & 0.801 & 0.482 & 0.414 & 0.468 & 0.458 & 0.490\\
Non-belief in the LLN & 0.722 & 0.498 & 0.538 & 0.592 & 0.601 & 0.671\\
Gambler's fallacy & 0.681 & 0.500 & 0.515 & 0.500 & 0.506 & 0.502\\
EGB --- loans & 0.601 & 0.483 & 0.476 & 0.494 & 0.495 & 0.501\\
EGB --- assets & 0.751 & 0.502 & 0.614 & 0.575 & 0.586 & 0.694\\
Limited attention & 0.700 & 0.502 & 0.582 & 0.563 & 0.566 & 0.574\\
\cmidrule(l){2-7}
Mean & 0.668 & 0.503 & 0.505 & 0.519 & 0.517 & 0.555\\
\quad Share of HUMAN & 1.00 & 0.02 & 0.03 & 0.11 & 0.10 & 0.33\\
\bottomrule
\end{tabular}
\end{table}

\textbf{Variance on the continuous scale.} Repeating the decomposition in \S\ref{sec:population} with
oriented, mid-rank tiles on the balanced sample ($n=518$) gives a human between-person share of
0.097, compared with 0.074 for the binary verdicts. Thus, binarization hides some person-level
variation in the human data. Relative to the human share, A5 falls from $1.05\times$ on binary
verdicts to $0.85\times$ on continuous scores; the other conditions range from $0.23\times$ to
$0.69\times$ on the continuous scale.

\section{Distribution of prediction gains across respondents}
\label{app:person-gains}

The mean gains in \S\ref{sec:ladder} do not show who benefits. To see that, we compare A5 with A1 and with A4 one
respondent at a time, using everyone with at least seven biases that have valid targets and predictions. Binary
accuracy is the share of matched wave-2 verdicts. Graded error is the mean absolute difference between model and
human, normalized by each bias's scale, over the 12 graded biases other than certainty preference, which has the
fewest valid answers.

Compared with A4, A5 lowers graded error for 59\% of respondents and raises it for 41\%, and the mean reduction is
0.015 (95\% interval 0.011--0.020). Compared with A1 the graded-error split is even, and the interval for the mean
includes zero. On binary accuracy, A5 is better than A1 for 46\% of respondents and worse for 33\%, with a mean gain
of 2.48 percentage points (95\% interval 1.45--3.49); over A4, the 95\% interval for its mean binary gain includes
zero. These person-weighted comparisons differ from the bias-weighted informedness in Table~\ref{tab:accuracy}. Since each response was generated once, we cannot assess whether a given respondent's
gain would persist across runs.

\begin{table}[H]
\centering\footnotesize
\caption{\textbf{Respondent-level changes under the behavioral-history condition (A5).} Better, Same, and
Worse are the percentages of respondents for whom A5 has lower, equal, or higher prediction error
than the comparison condition. Binary gains are percentage-point changes in accuracy across 14
biases; graded gains are reductions in normalized absolute error across 12 biases. Intervals are
95\% respondent-bootstrap intervals from 2,000 draws.}
\label{tab:person-gains}
\begin{tabular}{@{}llrrrrl@{}}
\toprule
Measure & A5 compared with & $n$ & Better & Same & Worse & Mean gain [95\% interval]\\
\midrule
Binary accuracy & A1 & 837 & 46.2 & 21.0 & 32.7 & 2.48 [1.45, 3.49]\\
Binary accuracy & A4 & 837 & 41.3 & 20.4 & 38.2 & 0.69 [$-0.32$, 1.62]\\
Graded error & A1 & 821 & 50.2 & 0.0 & 49.8 & 0.0036 [$-0.0010$, 0.0082]\\
Graded error & A4 & 821 & 59.2 & 0.0 & 40.8 & 0.0153 [0.0108, 0.0199]\\
\bottomrule
\end{tabular}
\end{table}

\section{Loss aversion by income and education}
\label{app:lossaversion}

In the human data, loss aversion increases modestly with income, and with education up to a
bachelor's degree, although adjacent intervals overlap (Table~\ref{tab:lossaversion}). Every condition with a persona profile (A2--A5) reverses this gradient. The reversal is
steepest under A2 and weakest under A5, but it remains present under A5. Under A1, the model almost
never rejects either gamble.

\begin{table}[H]
\centering\footnotesize
\caption{\textbf{Loss-aversion prevalence by income and education.} HUMAN is the wave-2 rate with
a 95\% bootstrap interval. Income and education groups are those used in
Figure~\ref{fig:columns} (Appendix~\ref{app:samples}). Overlapping human intervals mean that the human gradient should be treated as descriptive.}
\label{tab:lossaversion}
\setlength{\tabcolsep}{4.5pt}
\begin{tabular}{@{}llrcccccc@{}}
\toprule
Attribute & Group & $n$ & HUMAN & A1 & A2 & A3 & A4 & A5\\
\midrule
Income & low income & 328 & 0.555 [0.503, 0.607] & 0.009 & 0.826 & 0.869 & 0.820 & 0.509\\
 & middle income & 318 & 0.604 [0.547, 0.657] & 0.006 & 0.330 & 0.582 & 0.516 & 0.450\\
 & high income & 198 & 0.677 [0.611, 0.742] & 0.000 & 0.081 & 0.293 & 0.293 & 0.343\\
\midrule
Education & high school or less & 141 & 0.518 [0.440, 0.596] & 0.000 & 0.766 & 0.872 & 0.865 & 0.560\\
 & some college & 302 & 0.596 [0.540, 0.649] & 0.010 & 0.609 & 0.732 & 0.639 & 0.460\\
 & bachelor's & 230 & 0.643 [0.578, 0.704] & 0.004 & 0.309 & 0.522 & 0.483 & 0.448\\
 & postgraduate & 172 & 0.628 [0.558, 0.703] & 0.006 & 0.169 & 0.372 & 0.378 & 0.331\\
\bottomrule
\end{tabular}
\end{table}

\end{document}